\documentclass[pdflatex,sn-mathphys-num]{sn-jnl}

\usepackage{graphicx}%
\usepackage{multirow}%
\usepackage{amsmath,amssymb,amsfonts}%
\usepackage{amsthm}%
\usepackage{mathrsfs}%
\usepackage[title]{appendix}%
\usepackage{xcolor}%
\usepackage{textcomp}%
\usepackage{manyfoot}%
\usepackage{booktabs}%
\usepackage{algorithm}%
\usepackage{algorithmicx}%
\usepackage{algpseudocode}%
\usepackage{listings}%

\usepackage{amssymb}
\usepackage{amsmath}
\usepackage{graphicx}
\usepackage{subcaption}
\usepackage{booktabs}
\usepackage{tabularx}
\usepackage{adjustbox}
\usepackage{xspace}
\usepackage{multirow}
\usepackage{pifont}
\usepackage{xcolor}
\usepackage{url}
\usepackage{natbib}
\usepackage{longtable}
\usepackage{booktabs}
\usepackage{caption}
\usepackage{supertabular}
\usepackage{booktabs}
\usepackage{tabularray}

\newcommand{\method}{HiMS-Min\xspace}
\newcommand{\nmethod}{HiMS-Max\xspace}
\newcommand{\cmark}{\textcolor{green!80!black}{\ding{51}}}
\newcommand{\xmark}{\textcolor{red}{\ding{55}}}

\newcommand{\vericar}{{\sc Veri-Car}\xspace}

\theoremstyle{thmstyleone}%
\theoremstyle{thmstyletwo}%

\theoremstyle{thmstylethree}%

\begin{document}

\title[Article Title]{\vericar: Towards Open-world Vehicle Information Retrieval}


\author*[1]{\fnm{Andrés} \sur{Munoz}}\email{andres.munozgarza@jpmorgan.com}

\author[2]{\fnm{Nancy} \sur{Thomas}}\email{nancy.thomas@jpmchase.com}

\author[2]{\fnm{Annita} \sur{Vapsi}}\email{annita.vapsi@jpmchase.com}

\author[1]{\fnm{Daniel} \sur{Borrajo}}\email{daniel.borrajo@jpmchase.com}

\affil*[1]{\orgdiv{AI Research}, \orgname{JPMorgan Chase \& Co.}, \orgaddress{\city{Madrid}, \country{Spain}}}

\affil[2]{\orgdiv{AI Research}, \orgname{JPMorgan Chase \& Co.}, \orgaddress{\city{New, York}, \country{US}}}


\abstract{Many industrial and service sectors require tools to extract vehicle characteristics from images. This is a complex task not only by the variety of noise, and large number of classes, but also by the constant introduction of new vehicle models to the market. In this paper, we present \vericar, an information retrieval integrated approach designed to help on this task. It leverages supervised learning techniques to accurately identify the make, type, model, year, color, and license plate of cars. The approach also addresses the challenge of handling open-world problems, where new car models and variations frequently emerge, by employing a sophisticated combination of pre-trained models, and a hierarchical multi-similarity loss. \vericar demonstrates robust performance, achieving high precision and accuracy in classifying both seen and unseen data. Additionally, it integrates an ensemble license plate detection, and an OCR model to extract license plate numbers with impressive accuracy.}




\maketitle

\section{Introduction}
\label{sec:intro}

The ability to accurately verify vehicle information through images has significant practical applications. Examples appear in automated toll collection, insurance companies, car dealers, financial institutions, or enhanced security checks. \vericar is a state-of-the-art information retrieval approach developed to meet this need by providing precise visual inspection and verification of various vehicle attributes, including the make, model, type, year, color, and license plate.

The image retrieval problem tackled by \vericar is inherently open-world due to the constant introduction of new car models and modifications to existing ones. Traditional classifiers often fall short when confronted with out-of-distribution (OOD) inputs, such as car models or features that were not part of the training data. To overcome this, \vericar employs a combination of pre-trained models and metric learning, allowing it to generalize effectively even when encountering new and unseen data. \vericar labels new samples as OOD, and passes them to a human for labelling. Then, the new samples are added back into the database together with their label to allow for future classification of these examples. The retrieval and OOD detection models are general, and could be deployed in a diverse set of applications across several industries by training on datasets specific to these domains, such as retail products datasets.

The approach divides the retrieval task into two main components: (1) make, model, type, and year prediction, and (2) color prediction. Although both components could be merged into one, they were separated due to a lack of comprehensive datasets that include all attributes simultaneously.

For the model retrieval component, \vericar uses the Stanford Cars 196 Dataset~\cite{krause20133d}. The methodology involves leveraging the vast knowledge of pre-trained models, like CLIP~\cite{radford2021learning} and OpenCLIP~\cite{cherti2023reproducible}, and fine-tuning them through a metric loss that maps similar observations closer in the embedding space. The main objective is to generate high-quality embeddings, which can be used to classify seen and unseen classes. Given that car attributes have inherent hierarchical relationships to one another, we use a hierarchical version of multi-similarity loss. This approach not only captures complex relationships among the various vehicle attributes, but also prevents the model from making drastic mistakes during inference. 

In addition to the retrieval of the previous properties, \vericar includes a color retrieval component trained using the Kaggle Vehicle Color Recognition (VCoR) Dataset~\cite{panetta2021artificial}. Employing a similar architecture and multi-similarity loss, this model achieves excellent performance.

Furthermore, \vericar features a robust license plate retrieval approach. It integrates an ensemble of a YOLOv5 license plate detection model and a fine-tuned TrOCR license plate recognition model to extract license plate numbers from images with high accuracy~\cite{YOLOv5lp,trocrpaper, transformers, trocrhf}. The TrOCR base model was fine-tuned on images of synthetic license plates, which we generated to roughly resemble license plates from a range of countries and states. This allows the model to  generalize over countries rather than be country specific. On the Stanford Cars dataset, which contains license plates from a range of countries, \vericar's model significantly outperforms the baseline.

The main contributions of the paper include: (1) to the best of our knowledge, the first integration of an image-based vehicle information retrieval system capable of classifying: make, type, model, year, color and license plate in an open-world setting; (2) \method, a new hierarchical version of Multi-Similarity loss, that improves representations of unseen classes ;(3) a retrieval approach to color classification that improves performance over standard cross-entropy trained models; (4) a framework for training license plate recognition on purely synthetic data, that achieves high performance in the multinational domain relative to a country specific model; and (5) a system for generating synthetic data to train a license plate recognition model, either through pre-training and fine-tuning, or in a single training phase, which offers strong results in both the country specific and multinational domains, even when only a small amount of labeled country specific data is available.

Section~\ref{sec:relatedwork} presents related work and introduces the tasks of each component of the \vericar system.

\section{Related Work}
In this section, we will separate the related work on the three components of \vericar.
\label{sec:relatedwork}

\subsection{Vehicle Make and Model Recognition}

This is a well studied task in the closed-set setting~\cite{dai2017efficient,kemertas2020rankmi,zhang2022multi,lu2023efficient}. However, in recent years works on out-of-distribution detection and class discovery for open-world classification of vehicles have become increasingly popular. In these settings we are not only interested in creating a simple classifier, but rather include mechanisms to detect and deal with previously unseen classes. Wolf \textit{et al.}~\cite{wolf2024knowledge} leveraged a knowledge-distillation approach to include the probability of confusion between classes into the learning objective in order to improve OOD detection. Moreover, Vasquez-Santiago \textit{et al.}~\cite{vazquez2023vehicle} used a combination of a Gaussian Mixture Model and clustering for OOD detection and class discovery on a subset of 8 models of the VMMRdb database~\cite{tafazzoli2017large}. Furthermore, a vehicle model has inherent hierarchical information since a model is the combination of the maker, the car type (sedan, SUV, hatchback, etc.), the model name and the year it was released. Buzzelli \textit{et al.}~\cite{buzzelli2021revisiting} posed model recognition as a hierarchical classification task by adding a car type specific classification head and a make, model and year classification head for each on car type in the dataset. Contrary to previous works, our work poses the open-world problem as a retrieval task, in which \vericar not only classifies seen classes correctly, but also enforces system checks to identify previously unseen classes. At the same time, the hierarchy is not used to narrow down the classification task, but to create hierarchical clusters based on common characteristics on the vehicles.

\subsection{Vehicle Color Recognition}

This task has not seen as much attention as model recognition in the past. This is likely due to the strong challenges it poses in terms of data availability, ambiguity among colors, and class imbalance. While there have been attempts to solve the task using traditional computer vision techniques~\cite{hsieh2014vehicle,chen2014vehicle,brown2010example,tang2015vehicle}, the breadth of light and climate conditions that can be encountered, renders deep learning based methods most suitable~\cite{su2015vehicle,kim2024deep,hu2015vehicle}. Most color recognition methods based on traditional computer vision leveraged different color schemes like HSV, CIE lab, etc. according to the specific conditions in the dataset. Rachmadi \textit{et al.}~\cite{rachmadi2015vehicle} used a two-stream network to show that the RGB color space was better suited for the task compared to other traditionally used color spaces such as HSV or CIE Lab. Later on, Hu \textit{et al.}~\cite{hu2022joint} tried to lessen the effects of weather conditions jointly learning to classify vehicle colors, detection and an image denoising network that removed the noise introduced by rain on the image. SMNN-MSFF~\cite{hu2023vehicle} propose solving the problem of class imbalance through a smooth modulation loss on the color class. Although there have been works in the past using metric losses to solve the vehicle ReID problem~\cite{chu2019vehicle,shen2022joint,li2023clip},  to our knowledge, \vericar is the first work that tries to tackle vehicle color identification from the point-of-view of metric learning.

\subsection{License Plate Retrieval}

This is one of the most researched applications in computer vision, so it is considered, for the most part, a solved problem. Licence plate retrieval can be broken down into two tasks which can be performed independently or together: license plate detection (LPD) and license plate recognition (LPR). The former refers to the localization of a license plates within a larger image while the latter refers to the transcription of characters from the image of a license plate. Khan \textit{et al.}~\cite{LPsurvey} present a thorough review of LPR methods, including LPD methods. Here, we will focus on license plate retrieval methods which aim to provide general, rather than country specific, retrieval, but we encourage readers to examine Khan \textit{et al.}'s~\cite{LPsurvey} survey for a more comprehensive review of state of the art methods. They focus on neural network based methods since these tend to outperform classical methods, but note that license plate retrieval can be performed without the use of deep learning. For the same reason, we will also focus our review on deep learning based methods. Also, many of the best performing license plate retrieval methods are commercial products, and are therefore absent from the academic focused review we will discuss here. 

As discussed in~\cite{LPsurvey}, the lack of consistency in data makes the review of license plate retrieval methods more challenging. In particular, most works train and test on license plates from the same country or from countries with similar layouts. As demonstrated by Laroca \textit{et al.}~\cite{LPbias}, LPR models tend to perform very well on the dataset on which they were trained, but very poorly on other datasets. They showed this by testing the ability of a model trained on Chinese license plates to recognize Brazilian plates and vice versa. They found that cross-dataset performance was bad despite the fact that both plates use the Latin alphabet. This makes sense as license plate from different countries follow different templates, and, without exposure to a variety of styles during training, the model may be unable to recognize a range of plates. Among the methods presented in~\cite{LPsurvey}, very few of them include cross-dataset evaluation, and, furthermore, many of them are trained and evaluated on private datasets which are not available for testing. Moreover, the publicly available datasets represent a select few countries, making the task of generalized LPD very difficult.

One of the core goals of our license plate retrieval system is to provide multinational LPR without country specific customization. Henry \textit{et al.}~\cite{layoutdet} attempts to tackle this problem by including a layout detection module to classify license plates as single-line or double-line, which aids their character recognition and allows accurate retrieval for license plates from a wide variety of countries. This approach, however, involves training a separate module, and may not perform as well in a data-constrained setting. Additionally, there may be styles which are not as easily separable into single-line or double-line and therefore may be harder to detect. Laroca \textit{et al.}~\cite{laroca} take a similar approach, incorporating a unified LPD and layout classification stage, which classifies the license plate into one of five layout styles based on country of origin, and applies layout-specific post-processing rules based on the predicted style to improve recognition. Their system is limited to the detection of license plates from one of these five styles, and relies on the use of heuristics, which may make it less robust to a wide variety of license plates.

\vericar leverages synthetic license plate images to improve the generalizability of LPR models. The use of synthetic data for LPR is not new. Usmankhujaev~\cite{ksyth}  provides a tool to generate synthetic images of Korean license plates in a variety of styles, but does not tackle the use of synthetic data for broader LPR. Björklund \textit{et al.}~\cite{bjorklund} propose a system for generating synthetic license plates as well as a LPR model trained on these plates. They generate synthetic plates using country specific templates, which requires them to generate a new set of synthetic data, and train a new model for each country in order to obtain high accuracy. It also requires the presence of good templates for each license plate style. Our approach, on the other hand, generates license plates which do not follow the format of a particular country, and trains one model which provides good results on many plate styles without requiring retraining. Earl~\cite{mearl} provides a tool for synthetic license plate generation which, like ours, is not rooted in the style of any given country's license plate. Unlike ours, the license plates are placed on non-car images, and do not attempt to mimic a full car. Furthermore, the process for generated synthetic images is more simplistic, and contains less variety of licence plates styles than our approach.

As a baseline, we will use the method presented in~\cite{ccpd}, which is an end-to-end deep learning architecture called RPnet (Roadside Parking net) that performs both LPD and recognition simultaneously in a single forward pass. RPnet consists of a detection module with convolutional layers to predict the license plate bounding box, and a recognition module that uses ROI pooling to extract features from multiple layers for character recognition. The network is trained end-to-end by jointly optimizing localization and classification losses. They also introduce a large dataset called CCPD (Chinese City Parking Dataset) which we will use in our evaluation. The test dataset is split into seven categories which categorize the data: blur, rotate, challenge, fn (far/near), db (dark/bright), tilt, and green. The authors trained their model on this open source dataset, and made the model weights publicly available which, combined with the demonstrated state-of-the-art performance at the time of publication, make it a strong choice for a baseline model. 

\section{\vericar: System Overview}

\begin{figure*}[!t]
  \centering
   \includegraphics[width=\linewidth]{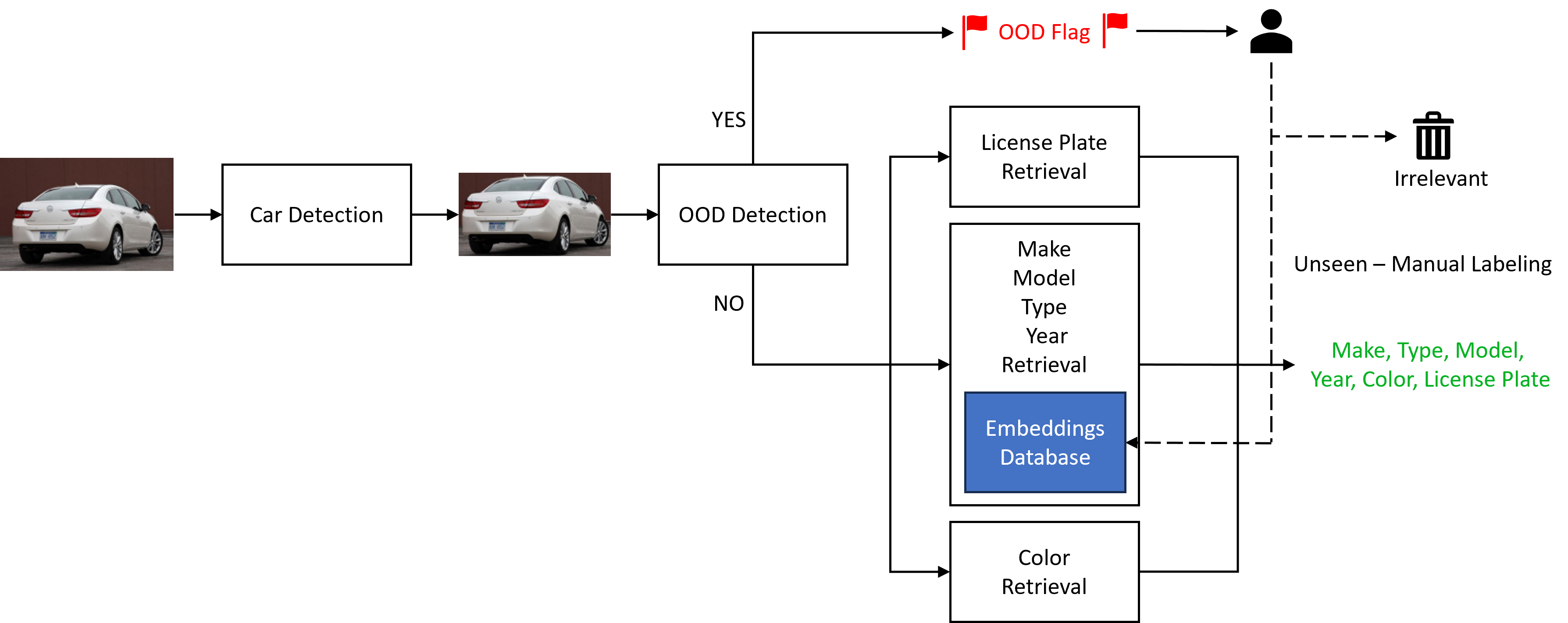}

   \caption{\vericar Architecture.}
   \label{fig:vericar_chart}
\end{figure*}

In this section we describe at a high level the components of the information retrieval approach. As mentioned in Section~\ref{sec:intro}, \vericar is composed of several independent modules. Figure~\ref{fig:vericar_chart} shows a high-level view of its architecture. During inference, a new car image is provided as input. First, the Car Detection module detects the bounding box of the car, and crops the image accordingly. The cropped image is then fed into the OOD Detection module, which is responsible for flagging if the image is considered in-distribution or out-of-distribution, i.e. if cars of similar attributes have been seen in the training data or not. In the case that the image is in-distribution, the License Plate Retrieval, Make, Model, Type, Year Retrieval and the Color Retrieval modules are responsible for extracting all necessary car attributes, namely, the license plate, make, model, type, year and color of the car. Conversely, if the image is flagged as out-of-distribution, a human-in-the-loop architecture is employed to review the image. If deemed relevant, the image is labeled and reintegrated into the \vericar's database. Otherwise, if the image is irrelevant — such as in cases where it does not contain a car —, it is discarded. The next sections describe in detail these modules.

\section{Make, Type, Model, Year \& Color Retrieval} 
In this section, we integrate both the Make, Type, Model, Year Retrieval and the Color Retrieval models. We discuss these two models together because of their similarities. We use $k$-NN as the main classification algorithm on the extracted embeddings of vehicle images as explained below in the Architecture subsection. At the same time, we introduce a hierarchical version of Multi-similarity loss~\cite{wang2020msloss} used to train the Make, Type, Model, Year Retrieval model while standard Multi-similarity loss~\cite{wang2020msloss} is used to train the Color Retrieval model. The only reason for keeping these two models distinct was the lack of comprehensive datasets combining all necessary labels.

\subsection{Multi-Similarity Loss}
The core of the model's retrieval process is based on metric learning, which aims to map similar observations closer in the embedding space. The multi-similarity loss was introduced by Wang \textit{et al.}~\cite{wang2020msloss} to enhance the performance of deep metric learning by effectively leveraging both the hardest-positive and hardest-negative pairs within each batch. This allows us to choose a positive and negative pair for each image in the batch with the most relevant information, optimizing the model to distinguish between similar and dissimilar samples effectively. The negative pair for an anchor image $i$ is defined as the image $k$, from a different class, with a higher similarity $s_{ik}$ than the hardest positive (image from the same class with lowest similarity). The positive pair is defined as the image $j$, belonging to the same class as $i$, with a lower similarity $s_{ij}$ than the hardest negative (image from a different class with the highest similarity). In the context of model retrieval, positive pairs consist of pairs of observations which belong to the same car make-type-model-year class, and negative pairs belong to two different car make-type-model-year classes. The formula for multi-similarity loss is shown in Equation~\ref{eq:multi_similarity_loss}, and ensures that the model learns to create clusters for samples of cars of the same characteristics, while maintaining a clear separation from samples of cars of different characteristics.

\begin{equation} 
\mathcal{L}_{MS} = \frac{1}{B} \sum_{i=1}^{B} \left[ \frac{1}{\alpha} \log \left( 1 + \sum_{j \in \mathcal{P}(i)} e^{-\alpha (s_{ij} - \lambda)} \right)\\ + \frac{1}{\beta} \log \left( 1 + \sum_{k \in \mathcal{N}(i)} e^{\beta (s_{ik} - \lambda)} \right) \right]
\label{eq:multi_similarity_loss} 
\end{equation}

where $B$ is the batch size, $\mathcal{P}(i)$ and $\mathcal{N}(i)$ denote the sets of positive and negative samples for the $i$-th anchor, respectively, $s_{ij}$ and $s_{ik}$ represent the similarity scores, and $\alpha$, $\beta$, and $\lambda$ are hyperparameters that control the scaling and margin of the loss. This formulation ensures that the model learns robust and discriminative feature representations by considering both positive and negative relationships within the data.

\subsection{Hierarchical Multi-Similarity Loss}
\label{sec:loss}

\begin{figure*}[!t]
  \centering
   \includegraphics[width=\linewidth]{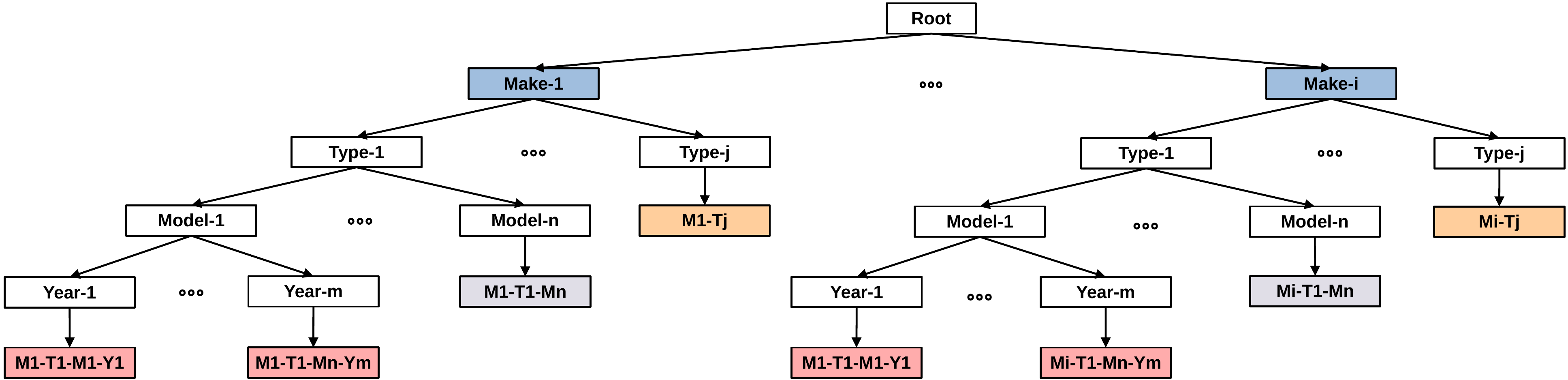}

   \caption{Cars labels hierarchies.}
   \label{fig:hierarchy}
\end{figure*}

Since the nature of our data is hierarchical, we created a hierarchical version of multi-similarity loss~\cite{wang2020msloss}.
A hierarchical multi-similarity loss identifies positive and negative pairs at all hierarchical levels for all samples in the batch aiming to create super-clusters of semantically similar samples. Formally, given an anchor image $i$, a set of hierarchy levels $L$ and a set of positive and negative pairs $Z_{l}(i) \, \forall\, l \in L$, where $Z_{l}(i) = P_{l}(i) \cup N_{l}(i)$, positive pairs $P_{l}(i)$ will be pushed closer the further down the hierarchy we go while negative pairs $N_{l}(i)$ will be pushed farther away as we move up in the hierarchy. Figure~\ref{fig:hierarchy} shows a visual representation of the hierarchical label structure created for the Make, Type, Model, Year Retrieval model. Samples at the leaves of the tree have the same make, type, model and year. This approach ensures that the model accurately learns the relationships among different vehicle attributes.

The hierarchical multi-similarity loss was created following Zhang \textit{et al.}'s ~\cite{zhang2022use} view on hierarchical contrastive learning where they use the inherent hierarchical tree-like relationships among the labels to optimize the model. Their proposed loss is shown in Equation~\ref{eq:hisupcon} and ensures that as we traverse the label tree downwards the loss at lower levels is never lower than at higher levels. The loss function is shown in Equation~\ref{eq:hisupcon}.

\begin{equation}
    \mathcal{L}_{HiSupCon} = \sum_{l=1}^{L} \frac{1}{|L|} \sum_{i=1}^{B} \frac{\lambda_{l}}{|P(i)|} \sum_{p_{l} \in \mathcal{P}_{l}} \text{max}(\mathcal{L}_{SupCon}(i, p^{i}_{l}), \mathcal{L}^{max}_{SupCon}(l-1))
    \label{eq:hisupcon}
\end{equation}

where $L$ is the set of hierarchical levels available in the taxonomy of the dataset, $B$ is the batch size, $P_l$ denotes the set of positive samples at hierarchical level $l$, $\mathcal{L}_{SupCon}$ is the standard SupCon loss proposed by ~\cite{khosla2020supervised}, while $\mathcal{L}^{max}_{SupCon}(l-1))$ is the max SupCon loss at the previous level of the hierarchy. $\lambda$ is a weighting factor, which is chosen to be $\lambda = \text{exp}(1/l)$. 

The above formulation forces lower levels of the hierarchy to have at least the same importance as that of higher levels. However, it might push sub-clusters too close to one another. Thus, we flipped the hierarchy, and forced higher levels of the hierarchy to have, at most, the same importance as lower levels. Such a formulation should create looser clusters as we go up the hierarchy, but still tight clusters at the leaves. Thus, the new loss function is defined in Equation~\ref{eq:hims}.

\begin{equation}
    \mathcal{L}_{HiMS-Min} = \sum_{L=l,...,1} \frac{1}{|L|} \sum_{i=1}^{B} \frac{\lambda_{l}}{|Z(i)|} \sum_{z_{l} \in \mathcal{Z}(i)_{l}} \text{min}(\mathcal{L}_{MS}(i, z^{i}_{l}), \mathcal{L}^{min}_{MS}(l+1)
    \label{eq:hims}
\end{equation}

where $L_{MS}$ is the Multi-similarity loss in Equation~\ref{eq:multi_similarity_loss} and $\lambda = \text{exp}(1/l)$.

Among the advantages of training with hierarchical loss is a better ability to generalize to unseen classes~\cite{zhang2022use}, and to provide enhanced partial answers to a user, even in cases where the model is unsure or makes mistakes at the fine-grain level. In our specific application, the trained model should in most cases be able to place a sample in the correct cluster for make and type.

\subsection{Make, Type, Model, Year Retrieval model}
In this subsection, we provide details on the dataset, architecture and training configurations of the Make, Type, Model, Year Retrieval model.

\subsubsection{Dataset}
The primary dataset used for the Make, Type, Model, Year Retrieval component is the Stanford Cars 196 Dataset~\cite{krause20133d}. This dataset provides a diverse collection of car images along with their associated labels for the make, model, vehicle type, and year. The dataset is composed of 16185 images and 196 classes. Each class is fully described by the concatenation of make, model, type and year labels. We divide the samples of each class in the dataset roughly in half. Then, we choose 160 classes as a seen set of classes which we use to train the model and 36 classes as an unseen set to simulate real-life open-world conditions. 

\subsubsection{Architecture}
We leverage pre-trained models, specifically OpenCLIP's ViT-B/16~\cite{dosovitskiy2020image} trained on LAION-2B~\cite{schuhmann2022laion} as the backbone to our model architecture. CLIP~\cite{radford2021learning} models are known for their ability to create compressed representations (embeddings) of images that capture intricate details and relationships. CLIP's learning objective maximizes the similarity between a vector representation of an image, and its corresponding text description in the embedding space. The combination of its training framework, and the size of the dataset make it a great initialization point.

In our experiments we use the image encoder and discard the text encoder provided in CLIP~\cite{radford2021learning}.  Then, a linear layer of size 128 is placed on top of the image encoder to learn a lower dimensional embedding to be used for $k$-NN classification. This architecture allows us to learn high-quality embedding representations of images with limited samples-per-class. 

\subsubsection{Model Configurations}

In order to properly asses our model and choice of loss function we benchmark our hierarchical multi-similarity loss against different versions of the loss and its non-hierarchical baseline. We tested the following configurations:
\begin{itemize}
    \item \textbf{MS:} Original Multi-similarity loss (Equation~\ref{eq:multi_similarity_loss}).
    \item \textbf{HiSupCon:} Original version of the loss as proposed by Zhang \textit{et al.}~\cite{zhang2022use}, as seen in Equation~\ref{eq:hisupcon}.
    \item \textbf{HiMS-Max:} Uses the top-down approach, as HiSupCon, but with Multi-Similarity loss as the base objective.
    \item \textbf{HiMS-Min (Ours):} Employs the bottom-up approach to the label hierarchy, using Multi-Similarity loss as the base objective, as described in Equation~\ref{eq:hims}.
\end{itemize}

\subsubsection{Training details}

The Stanford Cars dataset includes ground truth bounding box locations of the cars in the images, so we crop the car with a 5\% margin to ensure we do not leave important features out. After cropping, we resize the image to $256 \times 256$. Then, to increase the variety of samples, and prevent overfitting, we randomly crop an area of the image of at least 50\% and resize it to $224 \times 224$. Finally, we randomly flip the image horizontally. All configurations are trained with an Adam optimizer with a cyclic learning rate scheduling. The scheduler has a minimum learning rate of $2 \times 10^{-6}$ and maximum of $2 \times 10^{-4}$, we cycle through learning rates as in ~\cite{smith2017cyclical} for 200 epochs. We set the batch size to 600 for MS, \method and \nmethod and 350 for HiSupCon, due to memory restrictions. All experiments were run on 4 A10 GPUs.

\subsection{Color Retrieval}
\label{sec:color_ret}
In the following subsection, we give the details on the training configuration, architecture and dataset used to train the Color Retrieval model.

\subsubsection{Dataset}

As mentioned in previous sections, the color retrieval task was trained independently of the rest of the system since we found no open source datasets that included all relevant car characteristics. For this reason, the color label cannot be included in the hierarchy, and the Color Retrieval model is thus trained on the Kaggle Vehicle Color Recognition (VCoR)~\cite{panetta2021artificial} which includes 15 color classes and 10.5K images.

\subsubsection{Architecture}

Similar to the Make, Type, Model, Year retrieval model we leverage  OpenCLIP's ViT-B/16~\cite{dosovitskiy2020image} trained on LAION-2B~\cite{schuhmann2022laion} as the backbone to our architecture and add a 128 dimension linear head at the end to produce the embeddings.

\subsubsection{Model Configurations}

In order to properly asses our model and choice of loss function we benchmark Multi-Similarity loss against Cross Entropy and Rachmadi \textit{et al.}~\cite{rachmadi2015vehicle} Two-Stream Cross Entropy approach:
\begin{itemize}
    \item \textbf{MS:} Original Multi-similarity loss (Equation~\ref{eq:multi_similarity_loss}) with a linear layer of size 128 placed on top of the backbone.
    \item \textbf{Cross Entropy:} The linear layer is replaced by a single layer classifier.
    \item \textbf{Two-Stream:} Follows the approach outlined in~\cite{rachmadi2015vehicle}, but replaces the backbone with pretrained ViTs.
\end{itemize}

\subsubsection{Training Details}
As in the Make, Type, Model, Year Retrieval model, we resize the images to $256 \times 256$, randomly crop an area of the image of at least 66\%, and resize again it to $224 \times 224$. Finally, we randomly flip the image horizontally with a probability of 50\%. All models were trained with an Adam optimizer with a cyclic learning rate scheduling. The Multi-similarity model's scheduler has minimum learning rate of $2 \times 10^{-6}$ and maximum of $2 \times 10^{-4}$ and cycles through learning rates as in~\cite{smith2017cyclical} for 200 epochs. While the Cross Entropy model's cycles through a minimum of $5 \times 10^{-6}$ and maximum of $1 \times 10^{-4}$ for 50 epochs.

\subsection{Evaluation Metrics}
\label{sec:ret_inference}

\begin{figure*}[!t]
  \centering
   \includegraphics[width=\linewidth]{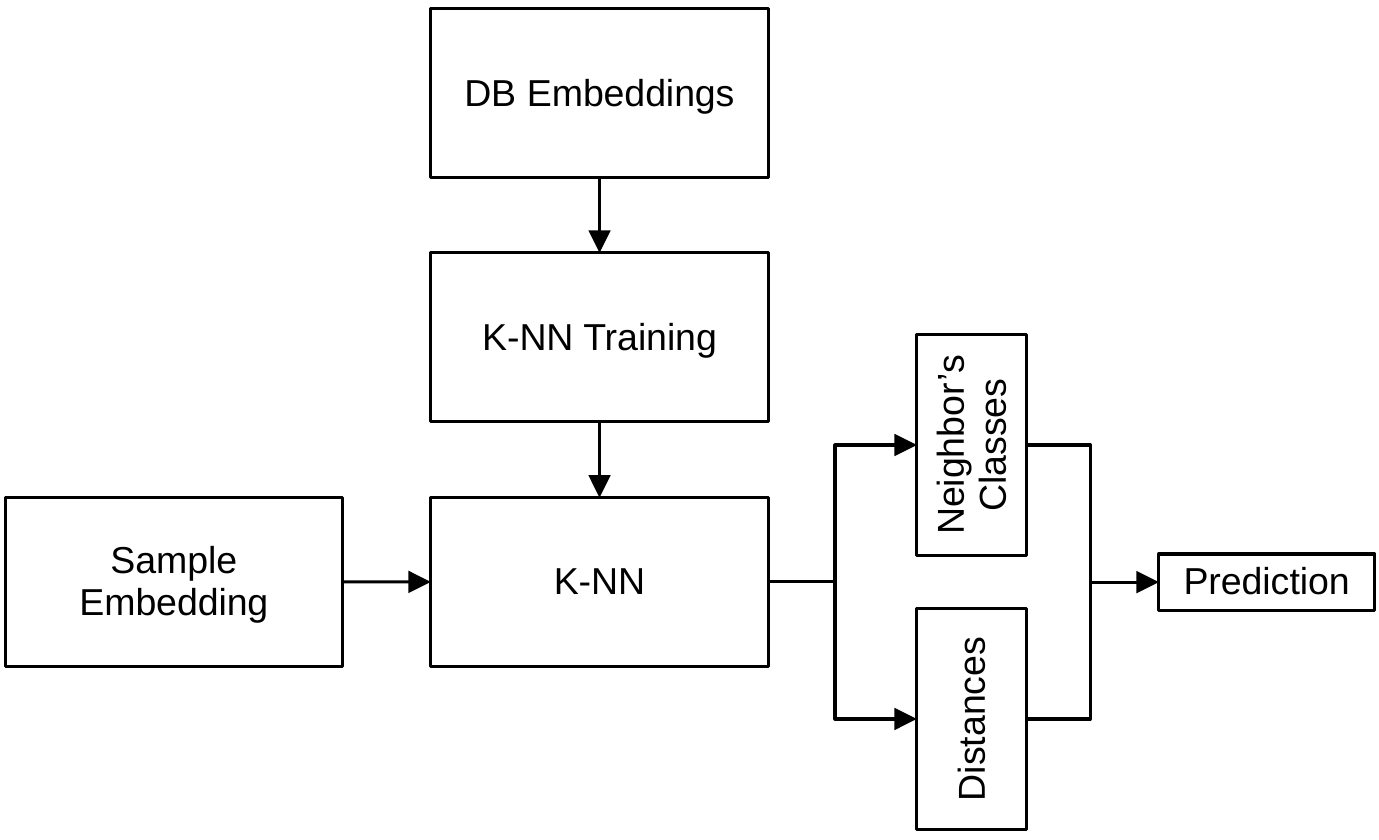}

   \caption{Retrieval flow chart.}
   \label{fig:retrieval_chart}
\end{figure*}

During inference, the Make, Type, Model, Year and Color Retrieval models use $k$-NN to assign a label to a sample embedding as illustrated in Figure~\ref{fig:retrieval_chart}. Embeddings for Make, Type, Model, Year Retrieval and Color Retrieval are extracted independently by their corresponding model. Distances are calculated using euclidean distance.

Since we are approaching the problem from a retrieval perspective, we will evaluate the model retrieval system using the following metrics:

\begin{itemize}
    
    \item \textbf{Precision@$k$}: 
    Precision for a specific value of $k$ in $k$-NN algorithm. In our experiments, we use $k=1$.

    \item \textbf{mAP@R}: Mean average precision at some recall level $R$.
\end{itemize}

Precision@$k$ measures the performance of the system by focusing on the first $k$ neighbors, while mAP@R measures the quality of the embedding space as a whole~\cite{musgrave2020metric}. All model configurations are tested at the make-type-model-year level of the hierarchy.

\subsection{Results}

The upcoming section lays-out the training details, presents and analyzes the results of the experiments on the Make, Type, Model, Year retrieval and color retrieval models.

\subsubsection{Make, Type, Model, Year Retrieval Model}
\begin{figure*}[!h]
  \centering
  \begin{subfigure}{0.497\linewidth}
     \includegraphics[width=0.98\linewidth]{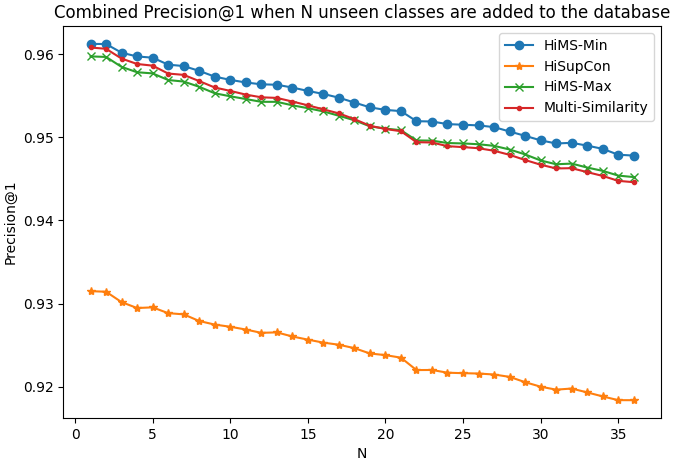}
    \caption{}
    \label{fig:unseen_prec}
  \end{subfigure}
  \hfill
  \begin{subfigure}{0.48\linewidth}
    \includegraphics[width=0.98\linewidth]{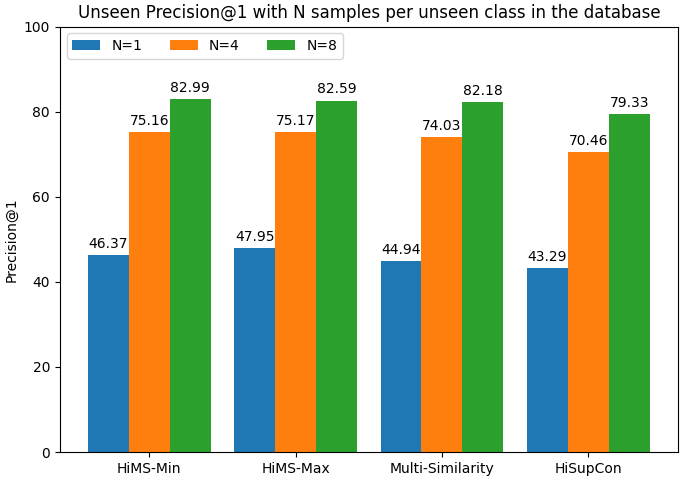}
    \caption{}
    \label{fig:select}
  \end{subfigure}
  \caption{(a) Change in Precision@1 when new classes are added into the retrieval database. (b) Precision@1 relative to number of samples per class in the database.}
  \label{fig:samples_per_class}
\end{figure*}

Table~\ref{tab:results1} shows the retrieval performance for Make, Type, Model, Year Model when only seen classes exist in the database. Results show that in this case \method outperforms baselines on Prec@1 while coming in second, behind Multi-similarity, on mAP@R. Table~\ref{tab:results2} shows similar results on a combined dataset that takes into account both seen and unseen classes. Results in Table~\ref{tab:results2} show that \method and Multi-Similarity can produce better separation of seen and unseen classes than other methods. Tables~\ref{tab:results3} and~\ref{tab:results4} show the breakdown of the combined performance found in Table~\ref{tab:results2}. Here we can see that \method beats most methods in both metrics in the unseen set, except \nmethod which ties it in Prec@1, while beating baselines on Prec@1 in the seen set. This points towards \method being able to find general features that distinguish between car models even if they were not seen during training. We believe these effects are caused by creating relationships at all levels of the hierarchy, but not inflating the importance at any one level. Furthermore, Figure~\ref{fig:unseen_prec} shows the progressive fall in the performance of the model as we add unseen classes to the database and test on them. The performance trend of \method and Multi-similarity point towards \method being more robust to classes being added into the database.

\begin{table*}[!htb]
    \caption{(a) Seen retrieval-based fine-grained classification results when only seen classes are in the database. (b) Combined retrieval-based fine-grained classification results. Combined refers to having both seen and unseen classes in the database and in the query set. (c) Seen retrieval-based fine-grained classification results, but with unseen classes in the database. (d) Unseen retrieval-based fine-grained classification results with seen classes in the database.}
    \begin{subtable}{.49\linewidth}
      \centering
        \begin{adjustbox}{width=0.99\linewidth}
            \begin{tabular}{l|c|c}
                \toprule
                Method & mAP@R & Prec@1 \\
                \midrule
                \textbf{HiSupCon} & 68.24 & 93.20 \\
                \textbf{Multi-Similarity} & \textbf{94.78} & 96.13 \\
                \textbf{\nmethod} & 91.65 & 96.03 \\
                \textbf{\method} & 94.11 & \textbf{96.18} \\
                \bottomrule
            \end{tabular}
        \end{adjustbox}
        \caption{Seen}
        \label{tab:results1}
    \end{subtable}%
    \begin{subtable}{.49\linewidth}
      \centering
        \begin{adjustbox}{width=0.99\linewidth}
            \begin{tabular}{l|c|c}
            \toprule
            Method & mAP@R & Prec@1 \\
            \midrule
            \textbf{HiSupCon} & 64.30 & 91.83 \\
            \textbf{Multi-Similarity} & \textbf{86.38} & 94.46 \\
            \textbf{\nmethod} & 83.60 & 94.52 \\
            \textbf{\method} & 86.02 & \textbf{94.78} \\
            \bottomrule
            \end{tabular}
        \end{adjustbox}
        \caption{Combined}
         \label{tab:results2}
    \end{subtable}
    \begin{subtable}{.49\linewidth}
      \centering
        \begin{adjustbox}{width=0.99\linewidth}
            \begin{tabular}{l|c|c}
            \toprule
            Method & mAP@R & Prec@1 \\
            \midrule
            \textbf{HiSupCon} & 65.18 & 91.77 \\
            \textbf{Multi-Similarity} & \textbf{92.03} & 95.11 \\
            \textbf{\nmethod} & 88.73 & 94.83 \\
            \textbf{\method} & 91.16 & \textbf{95.15} \\
            \bottomrule
            \end{tabular}
        \end{adjustbox}
        \caption{Seen with combined database}
         \label{tab:results3}
    \end{subtable}
    \begin{subtable}{.49\linewidth}
      \centering
        \begin{adjustbox}{width=0.99\linewidth}
            \begin{tabular}{l|c|c}
            \toprule
            Method & mAP@R & Prec@1 \\
            \midrule
            \textbf{HiSupCon} & 60.23 & 92.11 \\
            \textbf{Multi-Similarity} & 60.23 & 91.41 \\
            \textbf{\nmethod} & 59.92 & \textbf{93.08} \\
            \textbf{\method} & \textbf{62.30} & \textbf{93.08} \\
            \bottomrule
            \end{tabular}
        \end{adjustbox}
        \caption{Unseen with combined database}
         \label{tab:results4}
    \end{subtable}
\end{table*}

To properly asses the quality of the model, we study the minimum amount of samples per-class needed in the database to correctly classify an image from an unseen class. Therefore, we subsample our unseen classes, add them into the seen database, and then test retrieval of our unseen query set. Figure~\ref{fig:select} shows the average result over 100 runs. We can observe that using hierarchical supervision helps Multi-similarity trained models to be less sensitive to having less samples in the database, with \method scoring $82.99\%$ with 8 samples in the database. We believe this is due to the hard positive and negative mining employed by MS loss. Learning only through these samples at all levels reduces the bias introduced by easy, or redundant samples, helping the model focus on the distinctive features at all levels of hierarchy.

As explained in Section~\ref{sec:loss}, the model should be able to correctly assign labels at higher levels of the hierarchy to the samples it incorrectly predicted at the last level of the hierarchy. Table~\ref{tab:results5} shows that as we go up in the hierarchy, models get better at classifying samples that were incorrectly classified at the Year level (lowest level) of the hierarchy. Additionally, we can see that \nmethod and HiSupCon outperform \method. This was expected due to the lesser importance placed on creating relationships at higher hierarchical levels. Nonetheless, \method still makes a small gain, at the Make level of the hierarchy, over Multi-Similarity loss. Along with that, all models performed poorly at the Model level due to the dataset not having a diversity of years associated to each model.

\begin{table*}[!hb]
    \centering
    \caption{Accuracy at all hierarchy levels when retrieval fails at the make-type-model-year level.}
        \begin{adjustbox}{width=0.75\textwidth}
            \begin{tabular}{l|c|c|c}
            \toprule
            \multirow{2}{*}{Method} & \multicolumn{3}{c}{Prec@1} \\
            \cmidrule(lr){2-4}
            & Make-Type-Model & Make-Type & Make\\
            \midrule
            \textbf{HiSupCon} & 1.10 & 48.02 & 78.63\\
            \textbf{Multi-Similarity} & 1.16 & 48.84 & 77.13\\
            \textbf{\nmethod} & \textbf{1.51} & \textbf{52.08} & \textbf{78.87}\\
            \textbf{\method} & 1.18 & 46.67 & 77.25\\
            \bottomrule
            \end{tabular}
        \end{adjustbox}
        
        \label{tab:results5}
\end{table*}

\subsubsection{Color Retrieval Model}

\begin{figure*}[!hbt]
  \centering
   \includegraphics[width=0.66\linewidth]{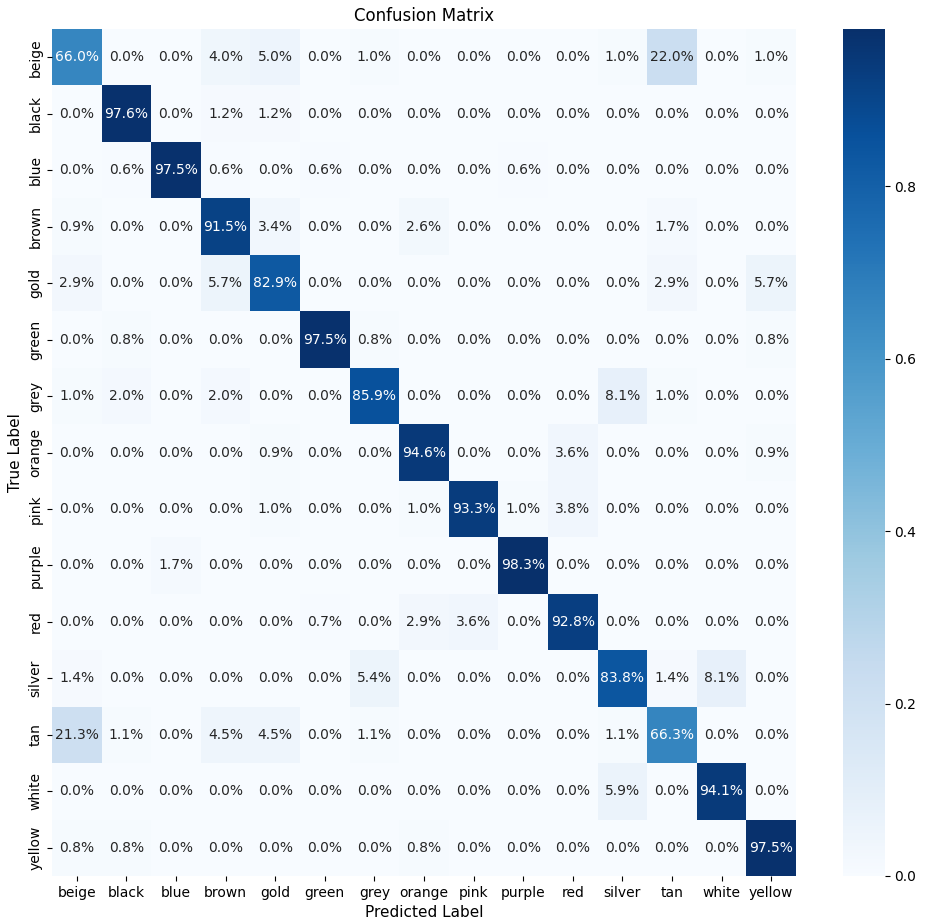}

   \caption{Multi-similarity loss color confusion matrix. Each cell reports the percentage of observations falling in the combination of predicted vs true labels.}
   \label{fig:confusion_matrix}
\end{figure*}

Table~\ref{tab:results_color} reports the performance of the Color Retrieval model on the Kaggle Vehicle Color Recognition (VCoR) Dataset. We perform well and are consistent across the tested performance metrics with Precision\@1 reaching 90.55\%, having better classification power than both baselines. Figure~\ref{fig:confusion_matrix} shows the confusion matrix for the Predicted versus True color labels. The errors of the model are reasonable, as similar colors tend to be confused. Specifically, note the combinations of beige and tan, as well as silver and grey.

\begin{table*}[!bth]
    \centering
    \caption{Color retrieval model results.}
        \begin{adjustbox}{width=0.7\textwidth}
            \begin{tabular}{l|c|c|c}
            \toprule
            \multirow{2}{*}{Method} & \multicolumn{3}{c}{Color}\\
            \cmidrule(lr){2-4}
            & mAP@R & Prec@1 & Top-1 Acc.\\
            \midrule
            \textbf{Multi-Similarity} & \textbf{89.68} & \textbf{90.55} & --\\
            \textbf{Cross Entropy} & 85.31 & 88.09 & 90.29\\
            \textbf{Two-Stream} & 84.29 & 87.53 & 89.52\\
            \bottomrule
            \end{tabular}
        \end{adjustbox}
        
        \label{tab:results_color}
\end{table*}

\section{OOD Detection}
\begin{figure*}[!tb]
  \centering
   \includegraphics[width=\linewidth]{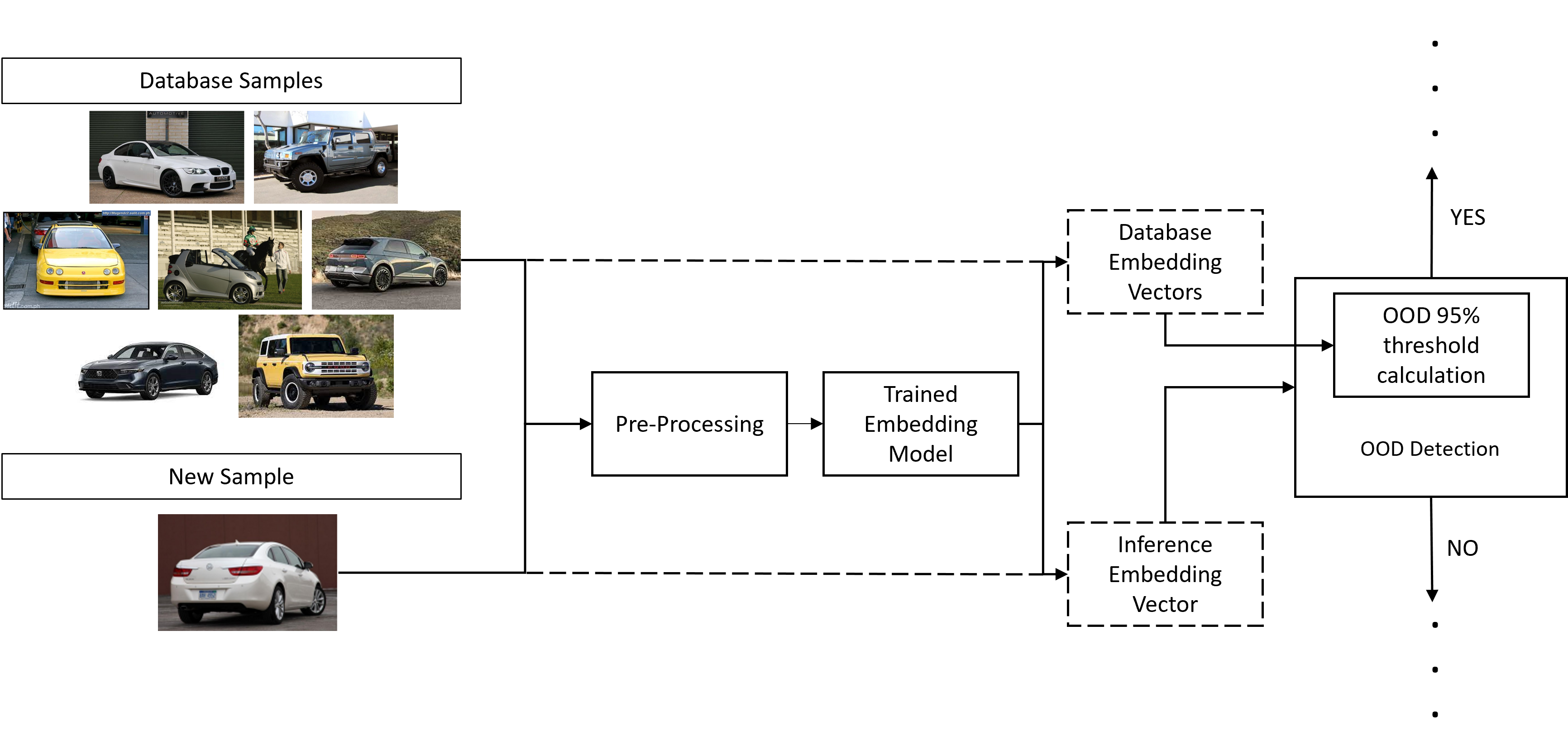}

   \caption{OOD detection flow chart.}
   \label{fig:ood_chart}
\end{figure*}

Out-of-Distribution (OOD) detection is a fundamental component of the system. We utilize KNN+~\cite{sun2022out}, an existing distance-based OOD detection method. KNN+ is a non-parametric approach that uses the distance of features in the feature space to their $k_{\text{th}}$ nearest neighbor to distinguish between in-distribution (ID) and OOD samples. Level set estimation on this metric determines the threshold above which a new sample is considered OOD. The method can be fine-tuned by searching for the $k_{\text{th}}$ neighbor's distance that achieves a 95\% recall of ID samples while minimizing the false positive rate (FPR) on OOD samples. Figure~\ref{fig:ood_chart} illustrates the inference flow of new samples within the OOD framework. Training set samples pass through the same pre-processing as for Make, Type, Model, Year Retrieval and the Trained Encoder Network to generate embedding vectors, thereby creating the in-distribution manifold in the embedding space. Similarly, a new image/sample is processed through the same modules to generate its embedding vector, which is then mapped to the embedding space.

\subsection{Evaluation Metrics}
We report our results with respect to two metrics: the false positive rate of OOD data when the TPR is 95\% (FPR95) and AUROC.

\subsection{Datasets}

We evaluate OOD detection on the 36 classes we held out as unseen in our OOD dataset.

\subsection{Results}

\begin{figure*}[!ht]
  \centering
   \includegraphics[width=0.85\linewidth]{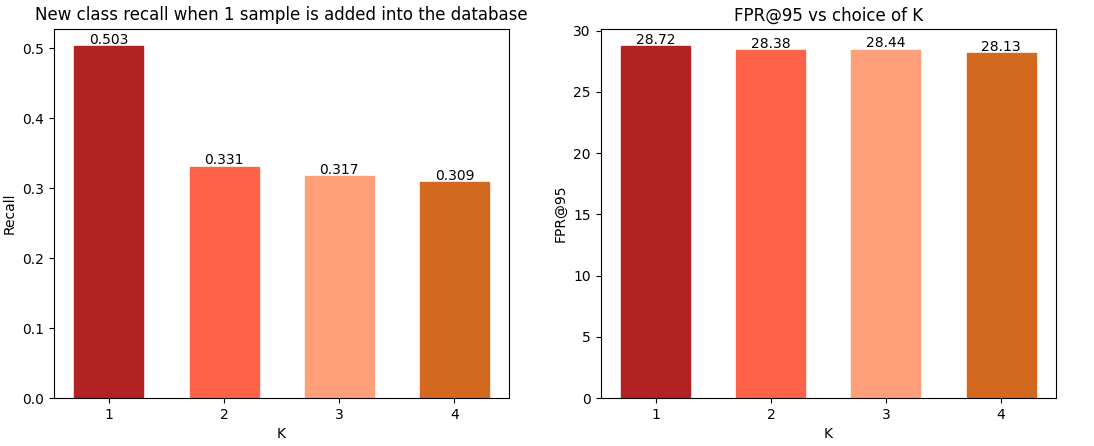}

   \caption{(Left) \method Recall of new class with a single samples in ID database with a different choice of K. (Right) \method FPR@95 performance under different choices of K.}
   \label{fig:kexp}
\end{figure*}

\begin{figure*}[!ht]
  \centering
   \includegraphics[width=0.65\linewidth]{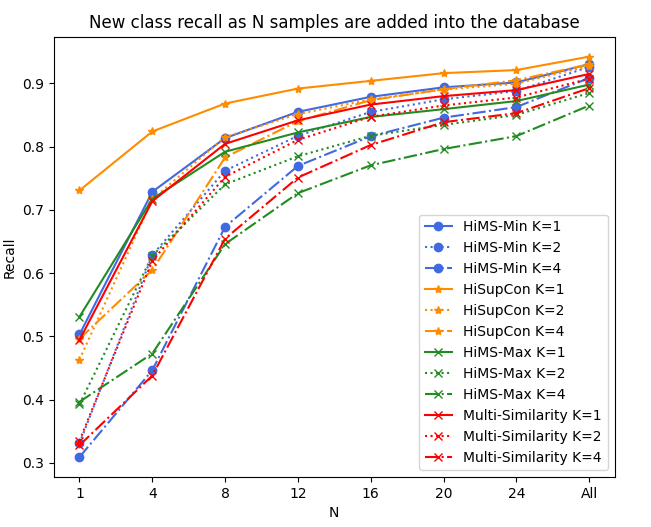}

   \caption{Recall of samples of a new class as samples are added into the database under different choices of K.}
   \label{fig:krecall}
\end{figure*}

\begin{figure*}[!ht]
  \centering
  \begin{subfigure}{0.657\linewidth}
    \includegraphics[width=\linewidth]{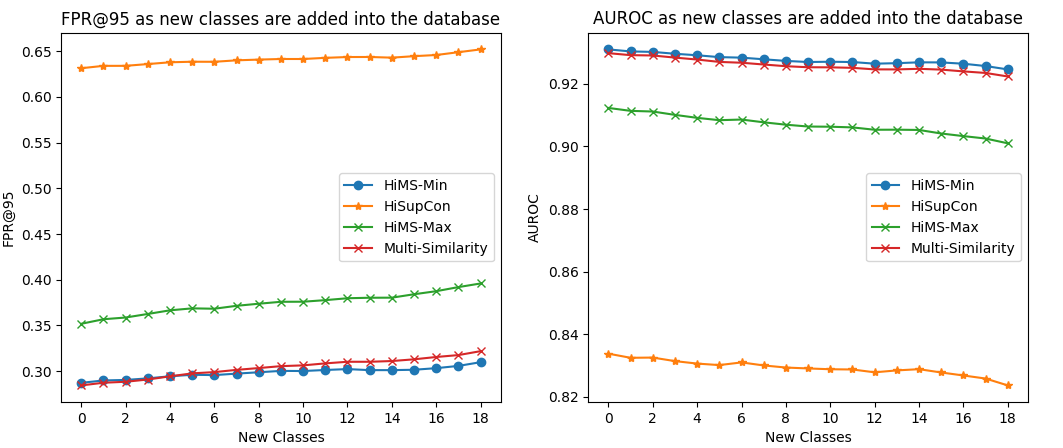}
    \caption{}
    \label{fig:add_ood}
  \end{subfigure}
  \hfill
  \begin{subfigure}{0.33\linewidth}
    \includegraphics[width=\linewidth]{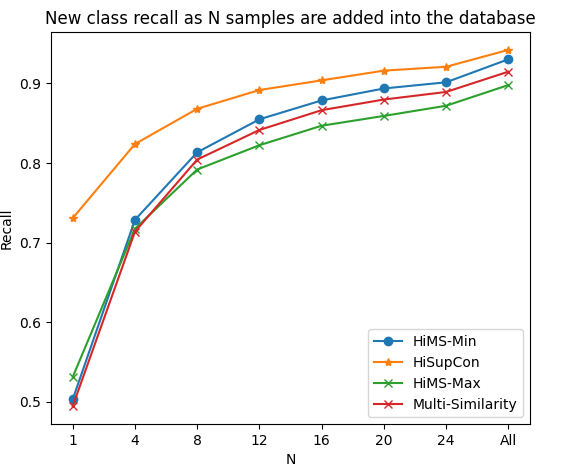}
    \caption{}
    \label{fig:add_samples_ood}
  \end{subfigure}
  
  \caption{(a) Effect of adding samples of a new class in FPR@95 and AUROC (b) Recall of samples of a new class as samples are added into the database.}
  \label{fig:ood_ablation}
\end{figure*}

Table~\ref{tab:ood_res} shows OOD results for models trained with hierarchical losses and Multi-similarity loss. For KNN+ experiments, we ran an exhaustive search for the best $k$ and landed on $k=1$. We not only want to optimize the FPR, but we would also like to find the minimum amount of samples needed to be added to the ID database in order to recognize new class samples as ID. The search took into account both FPR@95 and the average recall when a single sample of a new class is added to the ID database. While $k=1$ met both objectives for all methods, $k=4$ achieves the lowest FPR@95 for \method, and $k=1$ provided the best trade-off between average recall and FPR@95 (Figure~\ref{fig:kexp}). Sun \textit{et al.}~\cite{sun2022out} showed that datasets with a small number of samples per-class do not need to retrieve many neighbours to detect OOD samples. In our case, the Stanford Cars dataset averages 41 samples per class in the database, which is relatively small, and thus it makes sense that our chosen $k$ is lower than in other datasets. All methods perform better when using KNN+, which could be due to the lack of a sufficient number samples in the ID set to properly create a Gaussian ellipsoid, as required by the Mahalanobis distance~\cite{sehwag2021ssd}. Moreover, Figure~\ref{fig:kexp} shows that \method outperforms all baselines in FPR@95 if we were to choose $k=4$. Experiments in Figure~\ref{fig:krecall} not only shows a similar drop-off in performance when a single sample is added into the ID database, but a general decrease in performance as well for all methods. We believe this is due to, as the low mAP@R but high Prec@1 in Table~\ref{tab:results4} suggest, the quality of the clusters is not high leading to samples of unseen being more spread-out throughout the embedding space. Results shown are the average over 10 random runs for all 36 unseen classes. Furthermore, Figure~\ref{fig:add_samples_ood} shows a simplified version of Figure~\ref{fig:krecall}. We can observe that HiSupCon only needs 4 samples to classify $>80\%$ of the samples of the new class as ID, while \method and MS need 8 samples to achieve similar performance.

Finally, we test the performance of the OOD module while adding new classes into the database. Figure~\ref{fig:add_ood} shows the reduction in FPR@95 and AUROC performance, as we add new classes. In this experiment half of the classes were kept as unseen and the other half were added into the database one by one. We can observe that \method is more robust to new classes being added than other methods. The results provided are averaged over 100 runs. The order in which the 36 available classes are added into the database is randomized for each of the runs.

\begin{table*}[!htb]
    \centering
    \caption{OOD results.}
        \begin{adjustbox}{width=0.75\textwidth}
            \begin{tabular}{l|c|c|c|c}
            \toprule
            \multirow{2}{*}{Method} & \multicolumn{2}{c}{KNN+} & \multicolumn{2}{c}{Mahalanobis} \\
            \cmidrule(lr){2-3} \cmidrule(lr){4-5} 
            & FPR95 $\downarrow$ & AUROC $\uparrow$ & FPR95 $\downarrow$ & AUROC $\uparrow$ \\
            \midrule
            \textbf{HiSupCon} & 63.14 & 83.38 & 91.38 & 61.26 \\
            \textbf{Multi-Similarity} & \textbf{28.44} & 92.98 & 83.89 & 72.32 \\
            \textbf{\nmethod} & 35.18 & 91.23 & \textbf{81.79} & \textbf{73.97} \\
            \textbf{\method} & 28.72 & \textbf{93.10} & 85.47 & 70.80 \\
            \bottomrule
            \end{tabular}
        \end{adjustbox}
        \label{tab:ood_res}
\end{table*}

\section{License Plate Retrieval}

This section presents the License Plate retrieval component of \vericar, starting from its architecture.

\subsection{Architecture}
As discussed in Section~\ref{sec:relatedwork}, license plate retrieval can be split into two sub-tasks: LPD and LPR. \vericar takes a two stage approach, first performing LPD followed by LPR.

\begin{figure*}[!t]
  \centering
   \includegraphics[width=\linewidth]{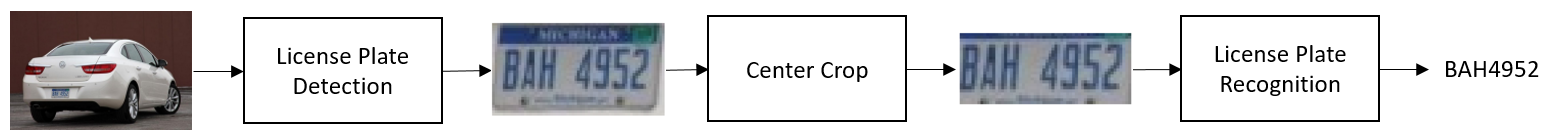}

   \caption{\vericar license retrieval framework.}
   \label{fig:lpdr_chart}
\end{figure*}

\subsection{License Plate Detection}
\vericar uses an open-source YOLOv5 model fine-tuned on the LPD task to detect and crop the license plates from car images Face~\cite{YOLOv5lp}. The model can be further fine-tuned for a specific dataset. However, in the interest of maintaining generalizability and due to the high accuracy of the original model, we decided to use the model out of the box.

\subsection{License Plate Recognition}
For LPR, \vericar uses an open-source TrOCR model. This model was first fine-tuned on the IAM handwriting dataset as it provided a good warm start relative to the pre-trained only model which was trained on a combination of textlines from PDF documents and synthetically generated handwritten text images~\cite{trocrpaper, transformers, trocrhf}. Additional fine-tuning was performed on 35771 generated synthetic license plate images. We will discuss the process for generating synthetic license plate images in Section~\ref{sec:lpr}.

As shown in Figure~\ref{fig:synth_plates}, many license plates contain text above or below the license plate. This text is not cropped out in the LPD module and since TrOCR is a single line OCR model, we need to first remove it from the license plate images or teach the model to ignore the top and bottom text if it is present. Initially, we tried the latter method but later found that center cropping allowed us to obtain good results without requiring an additional training module for the localization.

\subsection{End-to-end}
At inference time, an image of a car is first passed through the LPD module, which returns the bounding box of the license plate. This bounding box is used to crop the image of the car to just the image of the license plate. The license plate image is then passed into the LPR module. The LPD and recognition framework is shown in Figure~\ref{fig:lpdr_chart}.

\subsection{Dataset}
In this subsection we describe the datasets used for training and evaluation.

\subsubsection{License Plate Detection}
\begin{figure}[htbp]
    \centering
    \begin{subfigure}[b]{0.35\textwidth}
        \centering
        \includegraphics[width=\textwidth]{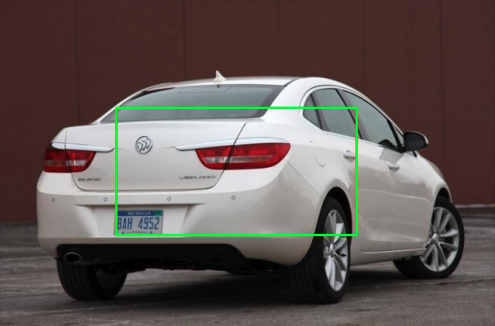}
        \caption{Ground Truth Label}
        \label{fig:subfig1}
    \end{subfigure}
    \hspace{0.05\textwidth} 
    \begin{subfigure}[b]{0.35\textwidth}
        \centering
        \includegraphics[width=\textwidth]{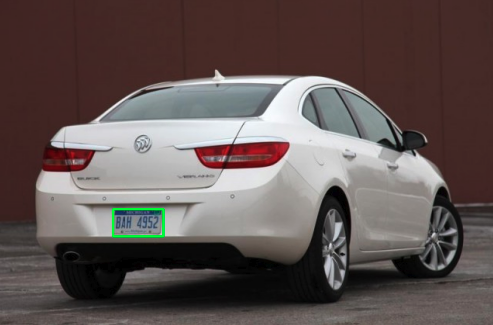}
        \caption{YOLOv5 Predicted Label}
        \label{fig:subfig2}
    \end{subfigure}
    \caption{Stanford Cars Dataset Bounding Box Misalignment}
    \label{fig:stan}
\end{figure}

The YOLOv5 model was trained on the Vehicle Registration Plates Data-set~\cite{YOLOv5data}. The LPD module is evaluated on the CCPD data as well as a custom dataset built by placing the synthetic plates in the license plate bounding boxes of the CCPD dataset. We do not evaluate LPD on the Stanford Cars dataset because only half of the images contain bounding box annotations. Furthermore, as illustrated in Figure~\ref{fig:stan}, the available annotations often over capture the license plate, leading to inaccurate evaluation scores.

\subsubsection{License Plate Recognition}
\label{sec:lpr}
The TrOCR model was pre-trained on a large-scale dataset of 684 million printed textline images generated from 2 million publicly available PDF documents, followed by task-specific datasets including 17.9 million synthetic handwritten textline images, 3.3 million printed textline images (including real-world receipts), and about 16 million scene text images from the MJSynth and SynthText datasets~\cite{trocrpaper, transformers, trocrhf}. The model we used was also fine-tuned on the IAM handwriting dataset~\cite{iam}. 

\begin{figure}[htbp]
    \centering
    \begin{minipage}{0.3\textwidth}
        \centering
        \includegraphics[width=\linewidth]{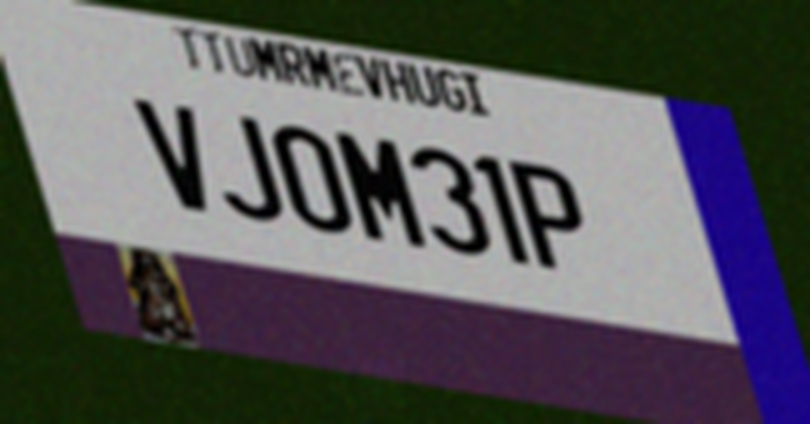}
    \end{minipage}\hfill
    \begin{minipage}{0.3\textwidth}
        \centering
        \includegraphics[width=\linewidth]{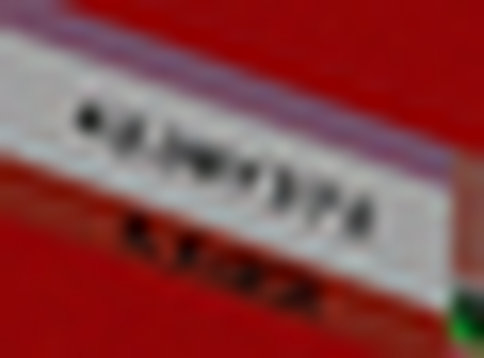}
    \end{minipage}\hfill
    \begin{minipage}{0.3\textwidth}
        \centering
        \includegraphics[width=\linewidth]{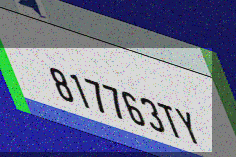}
    \end{minipage}

    \vspace{0.4cm} 

    \begin{minipage}{0.3\textwidth}
        \centering
        \includegraphics[width=\linewidth]{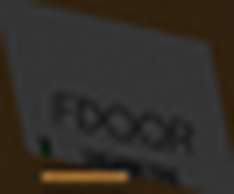}
    \end{minipage}\hfill
    \begin{minipage}{0.3\textwidth}
        \centering
        \includegraphics[width=\linewidth]{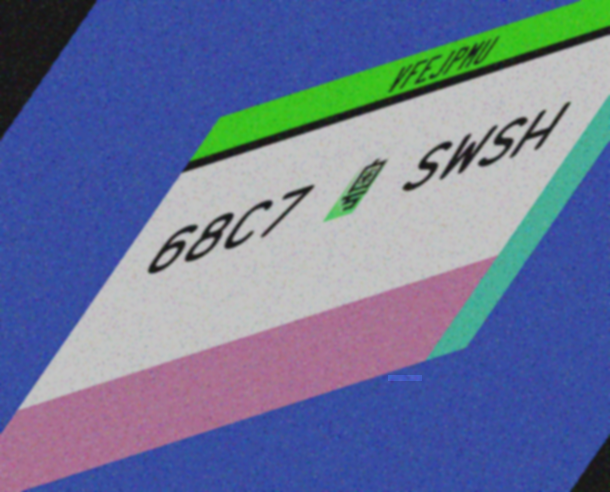}
    \end{minipage}\hfill
    \begin{minipage}{0.3\textwidth}
        \centering
        \includegraphics[width=\linewidth]{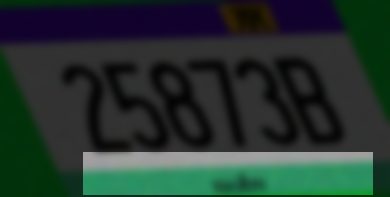}
    \end{minipage}

    \caption{Sample of synthetic license plates.}
    \label{fig:synth_plates}
\end{figure}

Additional fine-tuning was performed on a custom synthetic license plate dataset. We generated this dataset by creating rectangles which mimic license plates. It includes variations such as random blur, resolution reduction, tilt, size, color, shadow placement, presence of top text, presence of bottom text, and presence of icons to imitate real-world conditions from a variety of locations. These variations are applied with some probability where higher likelihoods are assigned to more typically observed cases. For example, dark or black text is more common, so these colors are given a higher probability in the synthetic data generation pipeline. Sample plates are shown in Figure~\ref{fig:synth_plates}. One may notice that the synthetic plates only vaguely resemble real license plates and do not conform to the style of any given country's license plates. This is because our goal is to build a model which generalizes well to the recognition of any license plate regardless of its origin country. The synthetic data generation process, however, could be easily tailored to generate plates that more closely match those of a particular region. When deploying the LPR model for a particular use case, we recommend additional training on such synthetic data or on real license plate examples from that region, if they exist.

\subsubsection{End-to-End}
To evaluate the whole license plate pipeline, we use three datasets: CCPD, the dataset of synthetic plates on CCPD cars, and the Stanford Cars dataset. Table~\ref{tab:data} contains a summary of the evaluation datasets, including an indication of which modules they are used to evaluate.

\subsection{Results}
We evaluate each of our two sub-tasks individually and also evaluate the pipeline as a whole. When evaluating the LPD, as well as the pipeline as a whole, we use RPnet as the baseline~\cite{ccpd}. For LPR, we will only present the results of our model. This is because RPnet is an end-to-end LDP and LPR model that expects images of a full car as input, whereas the data used for licence plate recognition evaluation contains only the license plate.

\subsubsection{License Plate Detection}
As the YOLOv5 model did not receive any fine-tuning, we report only the results of our LPD model here to inform the full pipeline results, as we will discuss later, but it will not serve as one of our main contributions. For LPD, we evaluate using two metrics: Dice Score (DS) (Equation~\ref{eq:ds}) and Intersection over Union (IoU) (Equation~\ref{eq:iou}). We report the results on the synthetic plates overlayed on the CCPD cars as well as the original images. 

\begin{equation}
    \text{DS}(X, Y) = \frac{2 \mid X \cap Y \mid}{\mid X \mid + \mid Y \mid}
    \label{eq:ds}
\end{equation}

\begin{equation}
    \text{IoU}(X, Y) = \frac{\mid X \cap Y \mid}{\mid X \cup Y \mid}
    \label{eq:iou}
\end{equation}

As we can see in Table~\ref{tab:lpd}, the YOLOv5 model outperforms RPnet on both the synthetic license plates overlaid on the CCPD cars and on the original CCPD cars. Moreover, the difference in performance between the two datasets for YOLOv5 is smaller than that of RPnet, suggesting that this model is more generalizable. CCPD seems to be overfit to CCPD training data and fails to capture many of the synthetic license plates. It is also worth noting that neither model has very high accuracy on average, so the LPD may present a weakness in the end-to-end evaluation.

\begin{table*}[!htb]
    \centering
    \caption{Datasets used for evaluation.}
        \begin{adjustbox}{width=\textwidth}
            \begin{tabular}{l|c|c|c|p{4.6cm}}
            Dataset & LPD & LPR & Full Pipeline & Description \\
            \midrule
            CCPD & \cmark & \xmark & \cmark & CCPD cars dataset (all subsets) \\
            CCPD Cropped License Plates & \xmark & \cmark & \xmark & CCPD cropped to just the license plate\\
            Synthetic License Plates & \xmark & \cmark & \xmark & Synthetically generated license plates \\
            Synthetic License Plates on CCPD & \cmark & \xmark & \cmark & Synthetically generated license plates overlaid on CCPD bounding box \\
            Stanford Cars & \xmark & \xmark & \cmark & Stanford cars dataset\\
            \bottomrule
            \end{tabular}
        \end{adjustbox}
        \label{tab:data}
\end{table*}

\begin{table*}[!htb]
    \centering
    \caption{Results of the license plate detection.}
    \begin{adjustbox}{width=0.75\textwidth}
        \begin{tabular}{l|c|c|c|c|c|c}
            \cmidrule{1-1} \cmidrule{2-3} \cmidrule{4-5} 
            \multirow{2}{*}{Dataset} & \multicolumn{2}{c|}{YOLOv5} & \multicolumn{2}{c|}{RPnet} \\
            \cmidrule{2-3} \cmidrule{4-5} 
            & DS $\uparrow$ & IoU $\uparrow$ & DS $\uparrow$ & IoU $\uparrow$ \\
            \cmidrule{1-1} \cmidrule{2-5} 
            Synthetic License Plates on CCPD Cars & \textbf{0.4638} & \textbf{0.4038} & 0.2537 & 0.1603 \\
            \cmidrule{1-1} \cmidrule{2-5} 
            CCPD &  \textbf{0.68276} & \textbf{0.6133} & 0.6273 & 0.4972 \\
            \cmidrule{1-1} \cmidrule{2-5} 
        \end{tabular}
    \end{adjustbox}
    \label{tab:lpd}
\end{table*}

\subsubsection{License Plate Recognition}
As previously discussed, we will not evaluate RPnet's performance on LPR alone and we will instead present the results of our LPR model relative to Tesseract, an open source OCR model~\cite{tesseractcode, tesseractpaper}. Tesseract was not trained specifically for LPR and is therefore expected to perform poorly on this task. We also introduce two new variants of our TrOCR model. The first is called TrOCR with CCPD, which is obtained by fine-tuning the original version of TrOCR trained on the synthetic images with a small sample (1\%) of the CCPD labeled images. The goal is to simulate a scenario where we have a small set of country specific labeled examples which may be used to fine-tune the multinational LPR model to achieve higher accuracy in the single country domain. The second is called TrOCR with CCPD + Synthetic, which has the same architecture as the original TrOCR model but is trained on the full set of synthetic and CCPD images. The goal here is to build a model which is able to achieve strong multinational and country specific accuracy in the case where a large amount of country specific labeled data does exist. We evaluate LPR according to two metrics: the percentage of predictions which exactly match the ground truth and the character error rate (CER) (Equation~\ref{eq:cer}) which measures the percent of characters that were incorrectly predicted by identifying the number of errors of each type (substitution, deletion, and insertion).

\begin{equation}
    \text{CER}(X, Y) = \frac{S + D + I}{N}
    \label{eq:cer}
\end{equation}
\begin{align*}
    \text{where} \\
    S &= \text{Number of substitutions} \\ 
    D &= \text{Number of deletions} \\ 
    I &= \text{Number of insertions} \\ 
    N &= \text{Total number of characters in the reference text}
\end{align*}

The results in terms of percentage correct and CER are shown in Tables~\ref{tab:lpr_percent_correct} and~\ref{tab:lpr_cer} respectively. As expected, the fine-tuned TrOCR model performs well on the synthetic license plates and less well on the CCPD license plates, while Tesseract fails on both. It is worth noting that in some subsets of the CCPD data, the model trained only on synthetic data performs decently, demonstrating the power  of synthetic data. The TrOCR with CCPD model and the TrOCR with CCPD + Synthetic model perform very well on CCPD license plates, with the latter slightly edging out the former in most cases likely due to the larger amount of data. Given fine-tuning on just 1\% of the data, however, TrOCR with CCPD is able to achieve nearly comparable performance, demonstrating that pre-training with synthetic data is a viable option in cases where there are few country specific labels. The TrOCR with CCPD + Synthetic mdel also performs well on just the synthetic data while the performance of TrOCR with CCPD in this case begins to drop off, indicating that during the fine-tuning process the model begins to learn more country specific information. Finally, the results also show that an OCR without LPR specific training, such as Tesseract, fails in the LPR setting.

\begin{table*}[!htb]
    \centering
    \caption{Results of the License Plate Recognition - Percent Correct.}
    \begin{adjustbox}{width=\textwidth}
        \begin{tabular}{p{2cm}|p{2cm}|p{2cm}|c|c|p{2cm}|c|c|c|c}
            \hline
            \textbf{Model} & Synthetic License Plates & CCPD License Plates & CCPD Blur & CCPD Rotate & CCPD Challenge & CCPD FN & CCPD DB & CCPD Tilt &CCPD Green \\
            \hline
            TrOCR & \textbf{73.34} & 5.86 & 0.06 & 2.32 & 2.10 & 6.89 & 20.23 & 11.91 &0.24 \\
            \hline
            TrOCR with CCPD & 59.70 & 76.75 & \textbf{66.01} & 88.70 & 74.06 & 81.56 & \textbf{70.33} & 84.62 & \textbf{61.57}\\
            \hline
            TrOCR with CCPD + Synthetic & 71.32 & \textbf{77.49} & 65.55 & \textbf{91.97} & \textbf{75.75} & \textbf{82.31} & 68.53 & \textbf{87.01} & 53.10\\
            \hline
            Tesseract & 0.00 & 0.00 & 0.00 & 0.00 & 0.00 & 0.00 & 0.00 &0.00 &0.00\\
            \hline
        \end{tabular}
    \end{adjustbox}
    \label{tab:lpr_percent_correct}
\end{table*}

\begin{table*}[!htb]
    \centering
    \caption{Results of the License Plate Recognition - CER}
    \begin{adjustbox}{width=\textwidth}
        \begin{tabular}{p{2cm}|p{2cm}|p{2cm}|c|c|p{2cm}|c|c|c|c}
            \hline
            \textbf{Model} & Synthetic License Plates & CCPD License Plates & CCPD Blur & CCPD Rotate & CCPD Challenge & CCPD FN & CCPD DB & CCPD Tilt &CCPD Green \\
            \hline
            TrOCR & \textbf{0.0837} & 0.5496 & 0.7916 & 0.6411 & 0.6263 & 0.4955 & 0.2693 & 0.3438 & 0.4819 \\
            \hline
            TrOCR with CCPD & 0.1377 & \textbf{0.0524} & \textbf{0.0781} & 0.0232 & \textbf{0.0600} & 0.0399 & \textbf{0.0666} & \textbf{0.0304} & \textbf{0.1179}\\
            \hline
            TrOCR with CCPD + Synthetic & 0.0931 & 0.0525 & 0.0823 & \textbf{0.0173} & 0.0604 & \textbf{0.0397} & 0.0744 & \textbf{0.0261} & 0.1238\\
            \hline
            Tesseract & 1.0865 & 1.0097 & 1.0076 & 1.0091 & 1.0060 & 1.0143 & 1.1084 & 1.0119 & 1.0075\\
            \hline
        \end{tabular}
    \end{adjustbox}
    \label{tab:lpr_cer}
\end{table*}

\subsubsection{End-to-end Evaluation}

For the end-to-end license plate detection and recognition model, we again evaluate our model, including the additional two variants involving country specific training, relative to RPnet. In the case of LPR, RPnet's architecture prevents it from predicting anything other than six character license plates with a particular set of characters following a particular pattern of the Chinese plates on which it was trained. Therefore, we evaluate it against filtered subsets of the synthetic and Stanford Cars which contain only instances of those license plates following the pattern of characters it could feasibly predict.

We will evaluate the end-to-end pipeline according to the following metrics:

\begin{itemize}
    \item LPD Accuracy: Percent of observations in which the LPD yielded a bounding box with a DS of at least 0.5.
    \item LPR Accuracy (Given Correct LPD): The percent of observations with an exactly correct predicted license plate among observations with a correct LPD.
    \item  Avg. CER (Given Correct LPD): The average CER among observations with a correct LPD.
    \item LPR Accuracy: The percent of observations with an exactly correct predicted license plate.
    \item  Avg. CER: The average CER.
\end{itemize}

The final two metrics serve as an overall evaluation of the model as they are impacted by both the LPD and the LPR. Table~\ref{tab:e2e_lpr} shows LPR accuracy for all models. This metric is highly related to Avg. CER, and thus can be used to determine which model performs best of each dataset overall. A more detailed set of results containing all metrics is shown in Table~\ref{tab:e2e}.

\begin{table*}[!htb]
    \centering
    \caption{LPR Accuracy of the end-to-end pipeline.}
    \renewcommand{\arraystretch}{1.2} 
    \begin{adjustbox}{width=0.94\textwidth}
        \begin{tabular}{p{3.5cm}|c|c|c|c|c}
            \hline
            Model & CCPD Full & Synthetic Full & Synthetic Full Filtered & Stanford Cars & Stanford Cars Filtered \\
            \hline
            YOLOv5 + TrOCR & 1.48 & \textbf{39.48} & \textbf{40.46} & 40.00 & 50.00 \\
            \hline
            YOLOv5 + TrOCR with CCPD & 42.48 & 31.48 & 35.54 & 31.43 & 58.33 \\
            \hline
            YOLOv5 + TrOCR with CCPD + Synthetic & \textbf{46.07} & 39.18 & 40.44 & \textbf{47.62} & \textbf{75.00} \\
            \hline
            RPnet & 37.36 & 0.00 & 0.00 & 0.00 & 0.00 \\
            \hline
        \end{tabular}
    \end{adjustbox}
    \label{tab:e2e_lpr}
\end{table*}

\begin{table*}[!htb]
    \centering
    \caption{Results of the end-to-end pipeline.}
    \renewcommand{\arraystretch}{0.7}
    \begin{adjustbox}{width=0.94\textwidth}
        \begin{tabular}{p{1cm}|l|c|c|c|c|c}
            \toprule
            \multirow{2}{*}{\rotatebox[origin=c]{90}{}} & \multirow{2}{*}{Dataset} &  \multirow{2}{*}{LPD Accuracy  $\uparrow$} & \multirow{2}{*}{\shortstack{LPR Accuracy \\ (Given Correct LPD)} $\uparrow$} & \multirow{2}{*}{\shortstack{Avg. CER \\ (Given Correct LPD)} $\downarrow$} & \multirow{2}{*}{LPR Accuracy $\uparrow$} & \multirow{2}{*}{Avg. CER $\downarrow$} \\
            & & & & & & \\
            \midrule
            \multicolumn{7}{c}{\textbf{YOLOv5 + TrOCR}} \\
            \midrule
            \multirow{8}{*}{\rotatebox[origin=c]{90}{\shortstack{Single \\ Country}}} & CCPD Full & 77.90 & 1.89 & 0.6365 & 1.48 & 0.6402 \\
            & \hspace{1 cm}CCPD Blur & 70.10 & 0.05 & 0.8400 & 0.03 & 0.8408 \\
            & \hspace{1 cm}CCPD Rotate & 89.51 & 7.65 & 0.3783 & 6.85 & 0.3896 \\
            & \hspace{1 cm}CCPD Challenge & 89.26 & 1.08 & 0.7041 & 0.98 & 0.7053 \\
            & \hspace{1 cm}CCPD FN & 75.42 & 1.86 & 0.6177 & 1.42 & 0.6221 \\
            & \hspace{1 cm}CCPD DB & 56.62 & 1.08 & 0.7002 & 0.62 & 0.7045 \\
            & \hspace{1 cm}CCPD Tilt & 68.13 & 2.88 & 0.4765 & 1.97 & 0.4880 \\
            & \hspace{1 cm}CCPD Green & 93.83 & 0.09 & 0.4602 & 0.08 & 0.4639 \\
            \hline
            \multirow{4}{*}{\rotatebox[origin=c]{90}{\shortstack{Mulit- \\ National}}} & Synthetic Full & 57.21 & 69.01 & 0.1176 & 39.48 & 0.1963 \\
            & Synthetic Full Filtered & 59.46 & 68.22 & 0.1103 & 40.56 & 0.1735 \\
            & Stanford Cars & - & - & - & 40.00 & 0.5930 \\
            & Stanford Cars Filtered  & - & - & - & 50.00 & 0.0972 \\
            \midrule
            \multicolumn{7}{c}{\textbf{YOLOv5 + TrOCR with CCPD}} \\
            \midrule
             \multirow{8}{*}{\rotatebox[origin=c]{90}{\shortstack{Single \\ Country}}} & CCPD Full & 77.91 & 54.31 & 0.1141 & 42.48 & 0.1232  \\
            & \hspace{1 cm}CCPD Blur & 70.10 & 49.25 & 0.1373 & 34.6 & 0.1444 \\
            & \hspace{1 cm}CCPD Rotate  & 89.51 & 55.35 & 0.1007 & 49.7 & 0.1087 \\
            & \hspace{1 cm}CCPD Challenge  & 89.26 & 58.13 & 0.1023 & 52.16 & 0.1079 \\
            & \hspace{1 cm}CCPD FN & 75.42 & 63.89 & 0.0904 & 48.4 & 0.1006 \\
            & \hspace{1 cm}CCPD DB & 56.62 & 50.62 & 0.1269 & 28.78 & 0.1404 \\
            & \hspace{1 cm}CCPD Tilt & 68.13 & 41.64 & 0.1454 & 28.4 & 0.1618 \\
            & \hspace{1 cm}CCPD Green & 93.83 & 64.62 & 0.0997 & 0.63 & 0.1057 \\
            \hline
            \multirow{4}{*}{\rotatebox[origin=c]{90}{\shortstack{Mulit- \\ National}}} & Synthetic Full & 57.21 & 55.03 & 0.1724 & 31.48 & 0.2494 \\
            & Synthetic Full Filtered & 59.46 & 59.77 & 0.1448 & 35.54 & 0.2038 \\
            & Stanford Cars & - & - & - & 31.43 & 0.6159 \\
            & Stanford Cars Filtered & - & - & - & 58.33 & 0.0972 \\
            \midrule
            \multicolumn{7}{c}{\textbf{YOLOv5 + TrOCR with CCPD + Synthetic}} \\
            \midrule
             \multirow{8}{*}{\rotatebox[origin=c]{90}{\shortstack{Single \\ Country}}} & CCPD Full & 77.91 & 58.93 & 0.1043 & 46.07 & 0.1148 \\
            & \hspace{1 cm}CCPD Blur & 70.10 & 51.16 & 0.1388 & 35.93 & 0.1473 \\
            &\hspace{1 cm}CCPD Rotate  & 89.51 & 65.74 & 0.0766 & 59.01 & 0.0862 \\
            &\hspace{1 cm}CCPD Challenge  &  89.26 & 62.29 & 0.0949 & 55.89 & 0.1014 \\
            &\hspace{1 cm}CCPD FN & 75.42 & 66.88 & 0.0841 & 50.64 & 0.0957 \\
            &\hspace{1 cm}CCPD DB & 56.62 & 51.70 & 0.1293 & 29.4 & 0.1448 \\
            &\hspace{1 cm}CCPD Tilt & 68.13  & 50.78 & 0.1204 & 34.63 & 0.1396 \\
            &\hspace{1 cm}CCPD Green & 93.83 & 52.18 & 0.1103 & 48.96 & 0.1170 \\
            \hline
            \multirow{4}{*}{\rotatebox[origin=c]{90}{\shortstack{Mulit- \\ National}}} & Synthetic Full &  57.21 & 68.48 & 0.1211 & 39.18 & 0.2052 \\
            & Synthetic Full Filtered & 59.46 & 68.02 & 0.1121 & 40.44 & 0.1770 \\
            & Stanford Cars  & - & - & - & 47.62 & 0.6008 \\
            & Stanford Cars Filtered & - & - & - & 75.00 & 0.0556  \\
            \midrule
            \multicolumn{7}{c}{\textbf{RPnet}} \\
            \midrule
             \multirow{8}{*}{\rotatebox[origin=c]{90}{\shortstack{Single \\ Country}}} & CCPD Full  & 95.93 & 38.77 & 0.2400 & 37.36 & 0.2569 \\
            & \hspace{1 cm}CCPD Blur  & 96.84 & 26.30 & 0.2790 & 25.57 & 0.2900 \\
            & \hspace{1 cm}CCPD Rotate  & 99.02 & 57.47 & 0.1425 & 57.07 & 0.1475 \\
            & \hspace{1 cm}CCPD Challenge  & 95.18 & 37.71 & 0.2353 & 36.17 & 0.2545 \\
            & \hspace{1 cm}CCPD FN  & 90.96 & 29.89 & 0.3224 & 27.39 & 0.3533 \\
            & \hspace{1 cm}CCPD DB  & 98.57 & 36.60 & 0.2366 & 36.21 & 0.2414 \\
            & \hspace{1 cm}CCPD Tilt  & 99.56 & 51.40 & 0.1701 & 51.19 & 0.1726 \\
            & \hspace{1 cm}CCPD Green  & 77.95 & 0.00 & 0.7421 & 0.00 & 0.7432 \\
            \hline
            \multirow{4}{*}{\rotatebox[origin=c]{90}{\shortstack{Mulit- \\ National}}} & Synthetic Full & 82.36 & 0.00 & 1.1276 & 0.00 & 1.1285 \\
            & Synthetic Full Filtered  & 82.51 & 0.00 & 0.9486 & 0.00 & 0.9489 \\
            & Stanford Cars  & - & - & - & 0.00 & 1.3074 \\
            & Stanford Cars Filtered  & - & - & - & 0.00 & 0.9306 \\
            \bottomrule
        \end{tabular}
    \end{adjustbox}
    \label{tab:e2e}
\end{table*}

The end-to-end results shown in Tables~\ref{tab:e2e_lpr} and~\ref{tab:e2e} depict RPnet's failure to capture license plates from countries on which it was not trained, even when the data has been filtered to include only license plates which fit the pattern of the training data. On the other hand, \vericar's YOLO+TrOCR model, which is fine-tuned on synthetic data only, offers stronger generalizability, outperforming RPnet by a wide margin on the Stanford Cars dataset which includes license plates from a range of countries. The results further underscore the effectiveness of \vericar's two additional TrOCR variants, both of which often surpass PRNet on the CCPD data. This shows that synthetic data can be valuable not only in a multinational context, but also when applied during pre-training for a country-specific detector. Similar to the LPR results, TrOCR with CCPD + Synthetic slightly outperforms TrOCR with CCPD, likely due to the increased volume of training data. Notably, within the green subset of CCPD plates, which represent images of green license plates while the training data contains only blue plates, TrOCR with CCPD significantly outperforms all other models. This suggests that the use of synthetic data is especially effective in accurately recognizing license plate numbers, despite the color variation in these plates.

Overall, our method demonstrates the ability for a model trained on simple and general synthetic license plates to perform multinational LPR well. It also shows that improved country specific license plate retrieval can be attained by training alongside synthetic data or by fine-tuning the model trained only on synthetic data. The approach is successful even in a data constrained setting. Importantly, training a country specific detector this way still allows for strong multinational detecting abilities, as demonstrated by the performance of TrOCR with CCPD and TrOCR with CCPD + Synthetic on the Stanford Cars dataset.

\section{Full System Evaluation}

In this section we evaluate the effectiveness of the system using a subset of 101 samples of the Stanford Cars dataset~\cite{Krause2013stanfrordcars}. We use the metrics already presented in each component of \vericar. As discussed in previous sections we struggled to find a training dataset with all required labels. To overcome this issue we use a dataset with complete labels for everything, but the car color. We asked three researchers to label the images, and used a voting system to decide the final labels. Our results are summarized in Table~\ref{tab:sys_res}. This new dataset includes 85 images of cars with labels seen during training and 16 images with unseen labels, never seen during training. The results in Table~\ref{tab:sys_res} summarize the performance of the each component of \vericar for all images corresponding to seen label instances. In this section we also present the results of the Out-of-Distribution detection model on the unseen set of images and summarize the overall performance of the system in correctly classifying all labels of an image on both the seen and unseen sets independently.

\begin{table*}[h!]
\centering
\caption{Performance of the developed models for color retrieval, Make, Type, Model, Year retrieval, and license plate recognition.}
\label{tab:results}
\begin{adjustbox}{width=0.8\textwidth}
\begin{tabular}{|l|c|c|}
\hline
\textbf{Model} & \textbf{Metric}& \textbf{Result}      \\ \hline
\multirow{2}{*}{\raggedright Color Retrieval}
& MAP@R & 75.08 \\
& Prec@1 & 78.14 \\ \hline
\multirow{2}{*}{\raggedright Color Retrieval (Popular Colors)} 
& MAP@R & 85.17 \\
& Prec@1 & 86.39 \\ \hline
\multirow{2}{*}{\raggedright Make, Type, Model, Year retrieval}    
& MAP@R & 83.43 \\
& Prec@1 & 88.21 \\ \hline
\multirow{2}{*}{\raggedright License Plate Recognition} & Percent Correct & 28.43 \\ 
& CER & 21.50 \\ \hline
\multirow{2}{*}{\raggedright License Plate Recognition (CER \textless 0.2)} 
& Percent Correct & 49.15 \\ 
& CER & 7.71 \\ \hline
\end{tabular}
\end{adjustbox}
\label{tab:sys_res}
\end{table*}

\subsection{Color Retrieval Model}
The Color Retrieval model achieved a MAP@R of 75.08 and a Prec@1 of 78.14. Through the labelling process, we observed that labeling certain colors can be ambiguous, particularly for less common colors like beige, tan and purple, potentially resulting in labeling errors. Recognizing the potential influence of less popular colors on model performance, we conducted an additional experiment focusing on only the popular color labels, including black, blue, grey, red, silver, and white. The results of this experiment showed a notable performance improvement of approximately 10\% in MAP@R and 8\% in Prec@1. Given the ambiguity in labeling these less common colors, along with the observed performance improvement on the subset of most popular colors, we conclude that the model's performance could be enhanced by training with additional images of underrepresented colors, if such data becomes available. Increasing the  number of images for these car colors would likely reduce the impact of labeling errors, allowing their distribution to reach a steady state. This, in turn, could help the model learn a more accurate representation of these colors and their relationship to similar hues in the dataset. At the same time, colors under different lighting conditions can look like other colors (e.g. brow, beige, tan and gold), so additional labels which include lighting conditions could further improve color recognition models.

\subsection{Make, Type, Model, Year Retrieval Model}

The model responsible for predicting car characteristics, including manufacturer, model, type, and class, was evaluated on the same test set. The model demonstrated robust performance, achieving a MAP@R of 83.43 and a Prec@1 of 88.21.

\subsection{License Plate Recognition Model}

The license plate detection model was evaluated using the percentage of correct predictions and the Character Error Rate (CER). The model's predictions exactly matched the ground truth in 28.43\% of the license plates with an average CER of 21.50\%. The majority of errors were due to omitted characters or confusion between visually similar characters, such as predicting a "1" instead of an "I". 

Acknowledging the challenges associated with character recognition using a model trained on solely synthetic data, we performed an additional experiment allowing for a 0.2 CER rate. Since most of the license plates in this test set have between 6 and 7 characters, a CER of 0.2 allows for between 1 and 2 character errors per license plate. Under this condition, the model's performance improved significantly, with 49.15\% of the predictions having a CER below the 0.2 cutoff. We anticipate further improvements in performance once the model is fine-tuned on the specific license plate layout of a particular country.

\subsection{OOD Detection}
The OOD detection model results are reported in Table~\ref{tab:ood_res_system}. Out of the 102 images in the test set, 16 were not seen by the model during training. The OOD detection model has an FPR95 OF 6.25\% and AUROC of 94.84\%, highlighting the success of the OOD detection model in flagging unseen samples. 

\begin{table*}[!htb]
    \centering
    \caption{OOD results.}
        \begin{adjustbox}{width=0.45\textwidth}
            \begin{tabular}{l|c|c}
            \toprule
            \multirow{2}{*}{Method} & \multicolumn{2}{c}{KNN+}\\
            \cmidrule(lr){2-3} 
            & FPR95 $\downarrow$ & AUROC $\uparrow$\\
            \midrule
            \textbf{\method} & 6.25 & 94.84\\
            \bottomrule
            \end{tabular}
        \end{adjustbox}
        \label{tab:ood_res_system}
\end{table*}

\subsection{System Accuracy}

In this section we examine the system's accuracy in the general classification problem. Our objective is to report the accuracy of the system in predicting all labels for each image (manufacturer, model, type, class, color and license plate). Table~\ref{tab:sys_acc} reports the accuracy of the system in classifying seen and unseen labels. We define the two problems as:
\begin{itemize}
    \item Building a database using all train set images with seen labels, and doing inference only on a test set of images with seen labels
    \item Building a database using all train set seen images and newly labeled (previously unseen) examples of images flagged as OOD by the OOD detector model. Finally, we inference only on images with unseen labels.
\end{itemize}
 In addition to the total accuracy of the system, we report accuracy when allowing for a 0.2 CER rate, and the accuracy when omitting the License Plate Recognition model results all together. As discussed previously, we expect the license plate model performance to change when fine tuned on the license plate layout of a particular country, which would eventually adjust the presented performance metrics. As such, we provide a lower limit (Total Accuracy) and upper limit (Total Accuracy no LPs) to the final expected performance.

\begin{table*}[!hb] 
\centering 
\caption{Total accuracy metrics for the whole system.} 
\label{tab:sys_acc} 
\begin{adjustbox}{width=0.98\textwidth}
\begin{tabular}{|c|c|c|c|} 
\hline 
\multirow{2}{*}{\textbf{Database Set / Query Set}} & \multirow{2}{*}{\textbf{Total Accuracy}} & \textbf{\textbf{Total Accuracy}} & \textbf{\textbf{Total Accuracy}} \\ 
&  & \textbf{with CER $<$0.2} & \textbf{no LPs}\\ \hline 
\textbf{Seen / Seen} & 30.23 & 52.33 & 90.70\\ 
\textbf{Seen \& Unseen  / Unseen} & 6.25 & 56.35 & 87.50 \\ \hline 
\end{tabular}
\end{adjustbox}
\end{table*} 

\section{Conclusion}
This paper introduces \vericar, a state-of-the-art open-world vehicle information retrieval system system. \vericar is able to accurately identify make, type, model,
year and color of not only models it was trained on, but up-to 36 additional models without retraining. Experiments show that a retrieval approach to color recognition achieves better results than the standard classification approach. \vericar also includes a license plate retrieval component, which leverages synthetic license plates to accurately detect, and recognize both single country and multinational plates with higher accuracy than the baseline model. Moreover, OOD detection results show FPR@95 robustness when new classes are added into the database. At the same time, experiments show that acquiring 8 samples per-new class is enough to not only accurately infer its characteristics, but to classify it as ID as well.



\section*{Disclaimer}

This paper was prepared for informational purposes by the Artificial Intelligence Research group of JPMorgan Chase \& Co. and its affiliates ("JP Morgan'') and is not a product of the Research Department of JP Morgan. JP Morgan makes no representation and warranty whatsoever and disclaims all liability, for the completeness, accuracy or reliability of the information contained herein. This document is not intended as investment research or investment advice, or a recommendation, offer or solicitation for the purchase or sale of any security, financial instrument, financial product or service, or to be used in any way for evaluating the merits of participating in any transaction, and shall not constitute a solicitation under any jurisdiction or to any person, if such solicitation under such jurisdiction or to such person would be unlawful.
 
© 2024 JPMorgan Chase \& Co. All rights reserved


\bibliography{sn-bibliography}


\begin{thebibliography}{51}
\ifx \bisbn   \undefined \def \bisbn  #1{ISBN #1}\fi
\ifx \binits  \undefined \def \binits#1{#1}\fi
\ifx \bauthor  \undefined \def \bauthor#1{#1}\fi
\ifx \batitle  \undefined \def \batitle#1{#1}\fi
\ifx \bjtitle  \undefined \def \bjtitle#1{#1}\fi
\ifx \bvolume  \undefined \def \bvolume#1{\textbf{#1}}\fi
\ifx \byear  \undefined \def \byear#1{#1}\fi
\ifx \bissue  \undefined \def \bissue#1{#1}\fi
\ifx \bfpage  \undefined \def \bfpage#1{#1}\fi
\ifx \blpage  \undefined \def \blpage #1{#1}\fi
\ifx \burl  \undefined \def \burl#1{\textsf{#1}}\fi
\ifx \doiurl  \undefined \def \doiurl#1{\url{https://doi.org/#1}}\fi
\ifx \betal  \undefined \def \betal{\textit{et al.}}\fi
\ifx \binstitute  \undefined \def \binstitute#1{#1}\fi
\ifx \binstitutionaled  \undefined \def \binstitutionaled#1{#1}\fi
\ifx \bctitle  \undefined \def \bctitle#1{#1}\fi
\ifx \beditor  \undefined \def \beditor#1{#1}\fi
\ifx \bpublisher  \undefined \def \bpublisher#1{#1}\fi
\ifx \bbtitle  \undefined \def \bbtitle#1{#1}\fi
\ifx \bedition  \undefined \def \bedition#1{#1}\fi
\ifx \bseriesno  \undefined \def \bseriesno#1{#1}\fi
\ifx \blocation  \undefined \def \blocation#1{#1}\fi
\ifx \bsertitle  \undefined \def \bsertitle#1{#1}\fi
\ifx \bsnm \undefined \def \bsnm#1{#1}\fi
\ifx \bsuffix \undefined \def \bsuffix#1{#1}\fi
\ifx \bparticle \undefined \def \bparticle#1{#1}\fi
\ifx \barticle \undefined \def \barticle#1{#1}\fi
\bibcommenthead
\ifx \bconfdate \undefined \def \bconfdate #1{#1}\fi
\ifx \botherref \undefined \def \botherref #1{#1}\fi
\ifx \url \undefined \def \url#1{\textsf{#1}}\fi
\ifx \bchapter \undefined \def \bchapter#1{#1}\fi
\ifx \bbook \undefined \def \bbook#1{#1}\fi
\ifx \bcomment \undefined \def \bcomment#1{#1}\fi
\ifx \oauthor \undefined \def \oauthor#1{#1}\fi
\ifx \citeauthoryear \undefined \def \citeauthoryear#1{#1}\fi
\ifx \endbibitem  \undefined \def \endbibitem {}\fi
\ifx \bconflocation  \undefined \def \bconflocation#1{#1}\fi
\ifx \arxivurl  \undefined \def \arxivurl#1{\textsf{#1}}\fi
\csname PreBibitemsHook\endcsname

\bibitem[\protect\citeauthoryear{Krause et~al.}{2013}]{krause20133d}
\begin{bchapter}
\bauthor{\bsnm{Krause}, \binits{J.}},
\bauthor{\bsnm{Stark}, \binits{M.}},
\bauthor{\bsnm{Deng}, \binits{J.}},
\bauthor{\bsnm{Fei-Fei}, \binits{L.}}:
\bctitle{3d object representations for fine-grained categorization}.
In: \bbtitle{Proceedings of the IEEE International Conference on Computer Vision Workshops},
pp. \bfpage{554}--\blpage{561}
(\byear{2013})
\end{bchapter}
\endbibitem

\bibitem[\protect\citeauthoryear{Radford et~al.}{2021}]{radford2021learning}
\begin{bchapter}
\bauthor{\bsnm{Radford}, \binits{A.}},
\bauthor{\bsnm{Kim}, \binits{J.W.}},
\bauthor{\bsnm{Hallacy}, \binits{C.}},
\bauthor{\bsnm{Ramesh}, \binits{A.}},
\bauthor{\bsnm{Goh}, \binits{G.}},
\bauthor{\bsnm{Agarwal}, \binits{S.}},
\bauthor{\bsnm{Sastry}, \binits{G.}},
\bauthor{\bsnm{Askell}, \binits{A.}},
\bauthor{\bsnm{Mishkin}, \binits{P.}},
\bauthor{\bsnm{Clark}, \binits{J.}}, \betal:
\bctitle{Learning transferable visual models from natural language supervision}.
In: \bbtitle{International Conference on Machine Learning},
pp. \bfpage{8748}--\blpage{8763}
(\byear{2021})
\end{bchapter}
\endbibitem

\bibitem[\protect\citeauthoryear{Cherti et~al.}{2023}]{cherti2023reproducible}
\begin{bchapter}
\bauthor{\bsnm{Cherti}, \binits{M.}},
\bauthor{\bsnm{Beaumont}, \binits{R.}},
\bauthor{\bsnm{Wightman}, \binits{R.}},
\bauthor{\bsnm{Wortsman}, \binits{M.}},
\bauthor{\bsnm{Ilharco}, \binits{G.}},
\bauthor{\bsnm{Gordon}, \binits{C.}},
\bauthor{\bsnm{Schuhmann}, \binits{C.}},
\bauthor{\bsnm{Schmidt}, \binits{L.}},
\bauthor{\bsnm{Jitsev}, \binits{J.}}:
\bctitle{Reproducible scaling laws for contrastive language-image learning}.
In: \bbtitle{Proceedings of the IEEE/CVF Conference on Computer Vision and Pattern Recognition},
pp. \bfpage{2818}--\blpage{2829}
(\byear{2023})
\end{bchapter}
\endbibitem

\bibitem[\protect\citeauthoryear{Panetta et~al.}{2021}]{panetta2021artificial}
\begin{barticle}
\bauthor{\bsnm{Panetta}, \binits{K.}},
\bauthor{\bsnm{Kezebou}, \binits{L.}},
\bauthor{\bsnm{Oludare}, \binits{V.}},
\bauthor{\bsnm{Intriligator}, \binits{J.}},
\bauthor{\bsnm{Agaian}, \binits{S.}}:
\batitle{Artificial intelligence for text-based vehicle search, recognition, and continuous localization in traffic videos}.
\bjtitle{AI}
\bvolume{2}(\bissue{4}),
\bfpage{684}--\blpage{704}
(\byear{2021})
\end{barticle}
\endbibitem

\bibitem[\protect\citeauthoryear{Face}{2023}]{YOLOv5lp}
\begin{botherref}
\oauthor{\bsnm{Face}, \binits{H.}}:
keremberke/yolov5m-license-plate.
\url{https://huggingface.co/keremberke/yolov5m-license-plate}
(2023)
\end{botherref}
\endbibitem

\bibitem[\protect\citeauthoryear{Li et~al.}{2021}]{trocrpaper}
\begin{botherref}
\oauthor{\bsnm{Li}, \binits{M.}},
\oauthor{\bsnm{Lv}, \binits{T.}},
\oauthor{\bsnm{Cui}, \binits{L.}},
\oauthor{\bsnm{Lu}, \binits{Y.}},
\oauthor{\bsnm{Florêncio}, \binits{D.A.F.}},
\oauthor{\bsnm{Zhang}, \binits{C.}},
\oauthor{\bsnm{Li}, \binits{Z.}},
\oauthor{\bsnm{Wei}, \binits{F.}}:
Trocr: Transformer-based optical character recognition with pre-trained models.
CoRR
\textbf{abs/2109.10282}
(2021).
\url{https://arxiv.org/abs/2109.10282}
\end{botherref}
\endbibitem

\bibitem[\protect\citeauthoryear{Wolf et~al.}{2020}]{transformers}
\begin{botherref}
\oauthor{\bsnm{Wolf}, \binits{T.}},
\oauthor{\bsnm{Debut}, \binits{L.}},
\oauthor{\bsnm{Sanh}, \binits{V.}},
\oauthor{\bsnm{Chaumond}, \binits{J.}},
\oauthor{\bsnm{Delangue}, \binits{C.}},
\oauthor{\bsnm{Moi}, \binits{A.}},
\oauthor{\bsnm{Cistac}, \binits{P.}},
\oauthor{\bsnm{Rault}, \binits{T.}},
\oauthor{\bsnm{Louf}, \binits{R.}},
\oauthor{\bsnm{Funtowicz}, \binits{M.}},
\oauthor{\bsnm{Davison}, \binits{J.}},
\oauthor{\bsnm{Shleifer}, \binits{S.}},
\oauthor{\bsnm{Platen}, \binits{P.}},
\oauthor{\bsnm{Ma}, \binits{C.}},
\oauthor{\bsnm{Jernite}, \binits{Y.}},
\oauthor{\bsnm{Plu}, \binits{J.}},
\oauthor{\bsnm{Xu}, \binits{C.}},
\oauthor{\bsnm{Le~Scao}, \binits{T.}},
\oauthor{\bsnm{Gugger}, \binits{S.}},
\oauthor{\bsnm{Drame}, \binits{M.}},
\oauthor{\bsnm{Lhoest}, \binits{Q.}},
\oauthor{\bsnm{Rush}, \binits{A.M.}}:
HuggingFace's Transformers: State-of-the-art Natural Language Processing.
\url{https://arxiv.org/abs/1910.03771}
(2020)
\end{botherref}
\endbibitem

\bibitem[\protect\citeauthoryear{Face}{2021}]{trocrhf}
\begin{botherref}
\oauthor{\bsnm{Face}, \binits{H.}}:
TrOCR.
\url{https://huggingface.co/docs/transformers/en/model_doc/trocr}
(2021)
\end{botherref}
\endbibitem

\bibitem[\protect\citeauthoryear{Dai et~al.}{2017}]{dai2017efficient}
\begin{bchapter}
\bauthor{\bsnm{Dai}, \binits{X.}},
\bauthor{\bsnm{Southall}, \binits{B.}},
\bauthor{\bsnm{Trinh}, \binits{N.}},
\bauthor{\bsnm{Matei}, \binits{B.}}:
\bctitle{Efficient fine-grained classification and part localization using one compact network}.
In: \bbtitle{Proceedings of the IEEE International Conference on Computer Vision Workshops},
pp. \bfpage{996}--\blpage{1004}
(\byear{2017})
\end{bchapter}
\endbibitem

\bibitem[\protect\citeauthoryear{Kemertas et~al.}{2020}]{kemertas2020rankmi}
\begin{bchapter}
\bauthor{\bsnm{Kemertas}, \binits{M.}},
\bauthor{\bsnm{Pishdad}, \binits{L.}},
\bauthor{\bsnm{Derpanis}, \binits{K.G.}},
\bauthor{\bsnm{Fazly}, \binits{A.}}:
\bctitle{Rankmi: A mutual information maximizing ranking loss}.
In: \bbtitle{Proceedings of the IEEE/CVF Conference on Computer Vision and Pattern Recognition},
pp. \bfpage{14362}--\blpage{14371}
(\byear{2020})
\end{bchapter}
\endbibitem

\bibitem[\protect\citeauthoryear{Zhang et~al.}{2022}]{zhang2022multi}
\begin{barticle}
\bauthor{\bsnm{Zhang}, \binits{H.}},
\bauthor{\bsnm{Li}, \binits{H.}},
\bauthor{\bsnm{Koniusz}, \binits{P.}}:
\batitle{Multi-level second-order few-shot learning}.
\bjtitle{IEEE Transactions on Multimedia}
\bvolume{25},
\bfpage{2111}--\blpage{2126}
(\byear{2022})
\end{barticle}
\endbibitem

\bibitem[\protect\citeauthoryear{Lu et~al.}{2023}]{lu2023efficient}
\begin{barticle}
\bauthor{\bsnm{Lu}, \binits{L.}},
\bauthor{\bsnm{Cai}, \binits{Y.}},
\bauthor{\bsnm{Huang}, \binits{H.}},
\bauthor{\bsnm{Wang}, \binits{P.}}:
\batitle{An efficient fine-grained vehicle recognition method based on part-level feature optimization}.
\bjtitle{Neurocomputing}
\bvolume{536},
\bfpage{40}--\blpage{49}
(\byear{2023})
\end{barticle}
\endbibitem

\bibitem[\protect\citeauthoryear{Wolf et~al.}{2024}]{wolf2024knowledge}
\begin{bchapter}
\bauthor{\bsnm{Wolf}, \binits{S.}},
\bauthor{\bsnm{Loran}, \binits{D.}},
\bauthor{\bsnm{Beyerer}, \binits{J.}}:
\bctitle{Knowledge-distillation-based label smoothing for fine-grained open-set vehicle recognition}.
In: \bbtitle{Proceedings of the IEEE/CVF Winter Conference on Applications of Computer Vision},
pp. \bfpage{330}--\blpage{340}
(\byear{2024})
\end{bchapter}
\endbibitem

\bibitem[\protect\citeauthoryear{Vázquez-Santiago et~al.}{2023}]{vazquez2023vehicle}
\begin{barticle}
\bauthor{\bsnm{Vázquez-Santiago}, \binits{D.-I.}},
\bauthor{\bsnm{Acosta-Mesa}, \binits{H.-G.}},
\bauthor{\bsnm{Mezura-Montes}, \binits{E.}}:
\batitle{Vehicle make and model recognition as an open-set recognition problem and new class discovery}.
\bjtitle{Mathematical and Computational Applications}
\bvolume{28}(\bissue{4}),
\bfpage{80}
(\byear{2023})
\end{barticle}
\endbibitem

\bibitem[\protect\citeauthoryear{Tafazzoli et~al.}{2017}]{tafazzoli2017large}
\begin{bchapter}
\bauthor{\bsnm{Tafazzoli}, \binits{F.}},
\bauthor{\bsnm{Frigui}, \binits{H.}},
\bauthor{\bsnm{Nishiyama}, \binits{K.}}:
\bctitle{A large and diverse dataset for improved vehicle make and model recognition}.
In: \bbtitle{Proceedings of the IEEE Conference on Computer Vision and Pattern Recognition Workshops},
pp. \bfpage{1}--\blpage{8}
(\byear{2017})
\end{bchapter}
\endbibitem

\bibitem[\protect\citeauthoryear{Buzzelli and Segantin}{2021}]{buzzelli2021revisiting}
\begin{barticle}
\bauthor{\bsnm{Buzzelli}, \binits{M.}},
\bauthor{\bsnm{Segantin}, \binits{L.}}:
\batitle{Revisiting the compcars dataset for hierarchical car classification: New annotations, experiments, and results}.
\bjtitle{Sensors}
\bvolume{21}(\bissue{2}),
\bfpage{596}
(\byear{2021})
\end{barticle}
\endbibitem

\bibitem[\protect\citeauthoryear{Hsieh et~al.}{2014}]{hsieh2014vehicle}
\begin{barticle}
\bauthor{\bsnm{Hsieh}, \binits{J.-W.}},
\bauthor{\bsnm{Chen}, \binits{L.-C.}},
\bauthor{\bsnm{Chen}, \binits{S.-Y.}},
\bauthor{\bsnm{Chen}, \binits{D.-Y.}},
\bauthor{\bsnm{Alghyaline}, \binits{S.}},
\bauthor{\bsnm{Chiang}, \binits{H.-F.}}:
\batitle{Vehicle color classification under different lighting conditions through color correction}.
\bjtitle{IEEE Sensors Journal}
\bvolume{15}(\bissue{2}),
\bfpage{971}--\blpage{983}
(\byear{2014})
\end{barticle}
\endbibitem

\bibitem[\protect\citeauthoryear{Chen et~al.}{2014}]{chen2014vehicle}
\begin{barticle}
\bauthor{\bsnm{Chen}, \binits{P.}},
\bauthor{\bsnm{Bai}, \binits{X.}},
\bauthor{\bsnm{Liu}, \binits{W.}}:
\batitle{Vehicle color recognition on urban road by feature context}.
\bjtitle{IEEE Transactions on Intelligent Transportation Systems}
\bvolume{15}(\bissue{5}),
\bfpage{2340}--\blpage{2346}
(\byear{2014})
\end{barticle}
\endbibitem

\bibitem[\protect\citeauthoryear{Brown}{2010}]{brown2010example}
\begin{bchapter}
\bauthor{\bsnm{Brown}, \binits{L.M.}}:
\bctitle{Example-based color vehicle retrieval for surveillance}.
In: \bbtitle{2010 7th IEEE International Conference on Advanced Video and Signal Based Surveillance},
pp. \bfpage{91}--\blpage{96}
(\byear{2010})
\end{bchapter}
\endbibitem

\bibitem[\protect\citeauthoryear{Tang and Xu}{2015}]{tang2015vehicle}
\begin{bchapter}
\bauthor{\bsnm{Tang}, \binits{Y.}},
\bauthor{\bsnm{Xu}, \binits{Y.}}:
\bctitle{Vehicle color recognition in static image for traffic enforcement camera system}.
In: \bbtitle{International Conference on Management, Computer and Education Informatization},
pp. \bfpage{310}--\blpage{313}
(\byear{2015})
\end{bchapter}
\endbibitem

\bibitem[\protect\citeauthoryear{Su et~al.}{2015}]{su2015vehicle}
\begin{bchapter}
\bauthor{\bsnm{Su}, \binits{B.}},
\bauthor{\bsnm{Shao}, \binits{J.}},
\bauthor{\bsnm{Zhou}, \binits{J.}},
\bauthor{\bsnm{Zhang}, \binits{X.}},
\bauthor{\bsnm{Mei}, \binits{L.}}:
\bctitle{Vehicle color recognition in the surveillance with deep convolutional neural networks}.
In: \bbtitle{2015 Joint International Mechanical, Electronic and Information Technology Conference (JIMET-15)},
pp. \bfpage{790}--\blpage{793}
(\byear{2015})
\end{bchapter}
\endbibitem

\bibitem[\protect\citeauthoryear{Kim}{2024}]{kim2024deep}
\begin{barticle}
\bauthor{\bsnm{Kim}, \binits{J.}}:
\batitle{Deep learning-based vehicle type and color classification to support safe autonomous driving}.
\bjtitle{Applied Sciences}
\bvolume{14}(\bissue{4}),
\bfpage{1600}
(\byear{2024})
\end{barticle}
\endbibitem

\bibitem[\protect\citeauthoryear{Hu et~al.}{2015}]{hu2015vehicle}
\begin{barticle}
\bauthor{\bsnm{Hu}, \binits{C.}},
\bauthor{\bsnm{Bai}, \binits{X.}},
\bauthor{\bsnm{Qi}, \binits{L.}},
\bauthor{\bsnm{Chen}, \binits{P.}},
\bauthor{\bsnm{Xue}, \binits{G.}},
\bauthor{\bsnm{Mei}, \binits{L.}}:
\batitle{Vehicle color recognition with spatial pyramid deep learning}.
\bjtitle{IEEE Transactions on Intelligent Transportation Systems}
\bvolume{16}(\bissue{5}),
\bfpage{2925}--\blpage{2934}
(\byear{2015})
\end{barticle}
\endbibitem

\bibitem[\protect\citeauthoryear{Rachmadi and Purnama}{2015}]{rachmadi2015vehicle}
\begin{botherref}
\oauthor{\bsnm{Rachmadi}, \binits{R.F.}},
\oauthor{\bsnm{Purnama}, \binits{I.}}:
Vehicle color recognition using convolutional neural network.
arXiv preprint arXiv:1510.07391
(2015)
\end{botherref}
\endbibitem

\bibitem[\protect\citeauthoryear{Hu et~al.}{2022}]{hu2022joint}
\begin{barticle}
\bauthor{\bsnm{Hu}, \binits{M.}},
\bauthor{\bsnm{Wu}, \binits{Y.}},
\bauthor{\bsnm{Fan}, \binits{J.}},
\bauthor{\bsnm{Jing}, \binits{B.}}:
\batitle{Joint semantic intelligent detection of vehicle color under rainy conditions}.
\bjtitle{Mathematics}
\bvolume{10}(\bissue{19}),
\bfpage{3512}
(\byear{2022})
\end{barticle}
\endbibitem

\bibitem[\protect\citeauthoryear{Hu et~al.}{2023}]{hu2023vehicle}
\begin{barticle}
\bauthor{\bsnm{Hu}, \binits{M.}},
\bauthor{\bsnm{Bai}, \binits{L.}},
\bauthor{\bsnm{Fan}, \binits{J.}},
\bauthor{\bsnm{Zhao}, \binits{S.}},
\bauthor{\bsnm{Chen}, \binits{E.}}:
\batitle{Vehicle color recognition based on smooth modulation neural network with multi-scale feature fusion}.
\bjtitle{Frontiers of Computer Science}
\bvolume{17}(\bissue{3}),
\bfpage{173321}
(\byear{2023})
\end{barticle}
\endbibitem

\bibitem[\protect\citeauthoryear{Chu et~al.}{2019}]{chu2019vehicle}
\begin{bchapter}
\bauthor{\bsnm{Chu}, \binits{R.}},
\bauthor{\bsnm{Sun}, \binits{Y.}},
\bauthor{\bsnm{Li}, \binits{Y.}},
\bauthor{\bsnm{Liu}, \binits{Z.}},
\bauthor{\bsnm{Zhang}, \binits{C.}},
\bauthor{\bsnm{Wei}, \binits{Y.}}:
\bctitle{Vehicle re-identification with viewpoint-aware metric learning}.
In: \bbtitle{Proceedings of the IEEE/CVF International Conference on Computer Vision},
pp. \bfpage{8282}--\blpage{8291}
(\byear{2019})
\end{bchapter}
\endbibitem

\bibitem[\protect\citeauthoryear{Shen et~al.}{2022}]{shen2022joint}
\begin{barticle}
\bauthor{\bsnm{Shen}, \binits{J.}},
\bauthor{\bsnm{Sun}, \binits{J.}},
\bauthor{\bsnm{Wang}, \binits{X.}},
\bauthor{\bsnm{Mao}, \binits{Z.}}:
\batitle{Joint metric learning of local and global features for vehicle re-identification}.
\bjtitle{Complex \& Intelligent Systems}
\bvolume{8}(\bissue{5}),
\bfpage{4005}--\blpage{4020}
(\byear{2022})
\end{barticle}
\endbibitem

\bibitem[\protect\citeauthoryear{Li et~al.}{2023}]{li2023clip}
\begin{bchapter}
\bauthor{\bsnm{Li}, \binits{S.}},
\bauthor{\bsnm{Sun}, \binits{L.}},
\bauthor{\bsnm{Li}, \binits{Q.}}:
\bctitle{Clip-reid: exploiting vision-language model for image re-identification without concrete text labels}.
In: \bbtitle{Proceedings of the AAAI Conference on Artificial Intelligence},
vol. \bseriesno{37},
pp. \bfpage{1405}--\blpage{1413}
(\byear{2023})
\end{bchapter}
\endbibitem

\bibitem[\protect\citeauthoryear{Khan et~al.}{2023}]{LPsurvey}
\begin{barticle}
\bauthor{\bsnm{Khan}, \binits{M.}},
\bauthor{\bsnm{Ilyas}, \binits{M.}},
\bauthor{\bsnm{Khan}, \binits{I.}},
\bauthor{\bsnm{Alshomrani}, \binits{S.}},
\bauthor{\bsnm{Rahardja}, \binits{S.}}:
\batitle{A review of license plate recognition methods employing neural networks}.
\bjtitle{IEEE Access}
\bvolume{PP},
\bfpage{1}--\blpage{1}
(\byear{2023})
\doiurl{10.1109/ACCESS.2023.3254365}
\end{barticle}
\endbibitem

\bibitem[\protect\citeauthoryear{Laroca et~al.}{2022}]{LPbias}
\begin{bchapter}
\bauthor{\bsnm{Laroca}, \binits{R.}},
\bauthor{\bsnm{Santos}, \binits{M.}},
\bauthor{\bsnm{Estevam}, \binits{V.}},
\bauthor{\bsnm{Luz}, \binits{E.}},
\bauthor{\bsnm{Menotti}, \binits{D.}}:
\bctitle{A first look at dataset bias in license plate recognition}.
In: \bbtitle{Proceedings of the SIBGRAPI},
pp. \bfpage{234}--\blpage{239}
(\byear{2022}).
\doiurl{10.1109/SIBGRAPI55357.2022.9991768}
\end{bchapter}
\endbibitem

\bibitem[\protect\citeauthoryear{Henry et~al.}{2020}]{layoutdet}
\begin{barticle}
\bauthor{\bsnm{Henry}, \binits{C.}},
\bauthor{\bsnm{Ahn}, \binits{S.Y.}},
\bauthor{\bsnm{Lee}, \binits{S.-W.}}:
\batitle{Multinational license plate recognition using generalized character sequence detection}.
\bjtitle{IEEE Access}
\bvolume{8},
\bfpage{35185}--\blpage{35199}
(\byear{2020})
\doiurl{10.1109/ACCESS.2020.2974973}
\end{barticle}
\endbibitem

\bibitem[\protect\citeauthoryear{Laroca et~al.}{2021}]{laroca}
\begin{barticle}
\bauthor{\bsnm{Laroca}, \binits{R.}},
\bauthor{\bsnm{Zanlorensi}, \binits{L.A.}},
\bauthor{\bsnm{Gonçalves}, \binits{G.R.}},
\bauthor{\bsnm{Todt}, \binits{E.}},
\bauthor{\bsnm{Schwartz}, \binits{W.R.}},
\bauthor{\bsnm{Menotti}, \binits{D.}}:
\batitle{An efficient and layout‐independent automatic license plate recognition system based on the yolo detector}.
\bjtitle{IET Intelligent Transport Systems}
\bvolume{15}(\bissue{4}),
\bfpage{483}--\blpage{503}
(\byear{2021})
\doiurl{10.1049/itr2.12030}
\end{barticle}
\endbibitem

\bibitem[\protect\citeauthoryear{Usmankhujaev}{2021}]{ksyth}
\begin{botherref}
\oauthor{\bsnm{Usmankhujaev}, \binits{S.}}:
Korean Car Plate Generator.
GitHub \url{https://github.com/Usmankhujaev/KoreanCarPlateGenerator}
(2021)
\end{botherref}
\endbibitem

\bibitem[\protect\citeauthoryear{Björklund et~al.}{2019}]{bjorklund}
\begin{barticle}
\bauthor{\bsnm{Björklund}, \binits{T.}},
\bauthor{\bsnm{Fiandrotti}, \binits{A.}},
\bauthor{\bsnm{Annarumma}, \binits{M.}},
\bauthor{\bsnm{Francini}, \binits{G.}},
\bauthor{\bsnm{Magli}, \binits{E.}}:
\batitle{Robust license plate recognition using neural networks trained on synthetic images}.
\bjtitle{Pattern Recognition}
\bvolume{93},
\bfpage{134}--\blpage{146}
(\byear{2019})
\doiurl{10.1016/j.patcog.2019.04.007}
\end{barticle}
\endbibitem

\bibitem[\protect\citeauthoryear{Earl}{2016}]{mearl}
\begin{botherref}
\oauthor{\bsnm{Earl}, \binits{M.}}:
Deep ANPR.
GitHub \url{https://github.com/matthewearl/deep-anpr}
(2016)
\end{botherref}
\endbibitem

\bibitem[\protect\citeauthoryear{Xu et~al.}{2018}]{ccpd}
\begin{bchapter}
\bauthor{\bsnm{Xu}, \binits{Z.}},
\bauthor{\bsnm{Yang}, \binits{W.}},
\bauthor{\bsnm{Meng}, \binits{A.}},
\bauthor{\bsnm{Lu}, \binits{N.}},
\bauthor{\bsnm{Huang}, \binits{H.}},
\bauthor{\bsnm{Ying}, \binits{C.}},
\bauthor{\bsnm{Huang}, \binits{L.}}:
\bctitle{Towards end-to-end license plate detection and recognition: A large dataset and baseline}.
In: \bbtitle{European Conference on Computer Vision}
(\byear{2018}).
\bcomment{\url{https://api.semanticscholar.org/CorpusID:52846606}}
\end{bchapter}
\endbibitem

\bibitem[\protect\citeauthoryear{Wang et~al.}{2020}]{wang2020msloss}
\begin{botherref}
\oauthor{\bsnm{Wang}, \binits{X.}},
\oauthor{\bsnm{Han}, \binits{X.}},
\oauthor{\bsnm{Huang}, \binits{W.}},
\oauthor{\bsnm{Dong}, \binits{D.}},
\oauthor{\bsnm{Scott}, \binits{M.R.}}:
Multi-Similarity Loss with General Pair Weighting for Deep Metric Learning.
\url{https://arxiv.org/abs/1904.06627}
(2020)
\end{botherref}
\endbibitem

\bibitem[\protect\citeauthoryear{Zhang et~al.}{2022}]{zhang2022use}
\begin{bchapter}
\bauthor{\bsnm{Zhang}, \binits{S.}},
\bauthor{\bsnm{Xu}, \binits{R.}},
\bauthor{\bsnm{Xiong}, \binits{C.}},
\bauthor{\bsnm{Ramaiah}, \binits{C.}}:
\bctitle{Use all the labels: A hierarchical multi-label contrastive learning framework}.
In: \bbtitle{Proceedings of the IEEE/CVF Conference on Computer Vision and Pattern Recognition},
pp. \bfpage{16660}--\blpage{16669}
(\byear{2022})
\end{bchapter}
\endbibitem

\bibitem[\protect\citeauthoryear{Khosla et~al.}{2020}]{khosla2020supervised}
\begin{barticle}
\bauthor{\bsnm{Khosla}, \binits{P.}},
\bauthor{\bsnm{Teterwak}, \binits{P.}},
\bauthor{\bsnm{Wang}, \binits{C.}},
\bauthor{\bsnm{Sarna}, \binits{A.}},
\bauthor{\bsnm{Tian}, \binits{Y.}},
\bauthor{\bsnm{Isola}, \binits{P.}},
\bauthor{\bsnm{Maschinot}, \binits{A.}},
\bauthor{\bsnm{Liu}, \binits{C.}},
\bauthor{\bsnm{Krishnan}, \binits{D.}}:
\batitle{Supervised contrastive learning}.
\bjtitle{Advances in Neural Information Processing Systems}
\bvolume{33},
\bfpage{18661}--\blpage{18673}
(\byear{2020})
\end{barticle}
\endbibitem

\bibitem[\protect\citeauthoryear{Dosovitskiy}{2020}]{dosovitskiy2020image}
\begin{botherref}
\oauthor{\bsnm{Dosovitskiy}, \binits{A.}}:
An image is worth 16x16 words: Transformers for image recognition at scale.
arXiv preprint arXiv:2010.11929
(2020)
\end{botherref}
\endbibitem

\bibitem[\protect\citeauthoryear{Schuhmann et~al.}{2022}]{schuhmann2022laion}
\begin{barticle}
\bauthor{\bsnm{Schuhmann}, \binits{C.}},
\bauthor{\bsnm{Beaumont}, \binits{R.}},
\bauthor{\bsnm{Vencu}, \binits{R.}},
\bauthor{\bsnm{Gordon}, \binits{C.}},
\bauthor{\bsnm{Wightman}, \binits{R.}},
\bauthor{\bsnm{Cherti}, \binits{M.}},
\bauthor{\bsnm{Coombes}, \binits{T.}},
\bauthor{\bsnm{Katta}, \binits{A.}},
\bauthor{\bsnm{Mullis}, \binits{C.}},
\bauthor{\bsnm{Wortsman}, \binits{M.}}, \betal:
\batitle{Laion-5b: An open large-scale dataset for training next generation image-text models}.
\bjtitle{Advances in Neural Information Processing Systems}
\bvolume{35},
\bfpage{25278}--\blpage{25294}
(\byear{2022})
\end{barticle}
\endbibitem

\bibitem[\protect\citeauthoryear{Smith}{2017}]{smith2017cyclical}
\begin{bchapter}
\bauthor{\bsnm{Smith}, \binits{L.N.}}:
\bctitle{Cyclical learning rates for training neural networks}.
In: \bbtitle{2017 IEEE Winter Conference on Applications of Computer Vision (WACV)},
pp. \bfpage{464}--\blpage{472}
(\byear{2017})
\end{bchapter}
\endbibitem

\bibitem[\protect\citeauthoryear{Musgrave et~al.}{2020}]{musgrave2020metric}
\begin{bchapter}
\bauthor{\bsnm{Musgrave}, \binits{K.}},
\bauthor{\bsnm{Belongie}, \binits{S.}},
\bauthor{\bsnm{Lim}, \binits{S.-N.}}:
\bctitle{A metric learning reality check}.
In: \bbtitle{Computer Vision--ECCV 2020: 16th European Conference, Glasgow, UK, August 23--28, 2020, Proceedings, Part XXV 16},
pp. \bfpage{681}--\blpage{699}
(\byear{2020})
\end{bchapter}
\endbibitem

\bibitem[\protect\citeauthoryear{Sun et~al.}{2022}]{sun2022out}
\begin{bchapter}
\bauthor{\bsnm{Sun}, \binits{Y.}},
\bauthor{\bsnm{Ming}, \binits{Y.}},
\bauthor{\bsnm{Zhu}, \binits{X.}},
\bauthor{\bsnm{Li}, \binits{Y.}}:
\bctitle{Out-of-distribution detection with deep nearest neighbors}.
In: \bbtitle{International Conference on Machine Learning},
pp. \bfpage{20827}--\blpage{20840}
(\byear{2022})
\end{bchapter}
\endbibitem

\bibitem[\protect\citeauthoryear{Sehwag et~al.}{2021}]{sehwag2021ssd}
\begin{botherref}
\oauthor{\bsnm{Sehwag}, \binits{V.}},
\oauthor{\bsnm{Chiang}, \binits{M.}},
\oauthor{\bsnm{Mittal}, \binits{P.}}:
Ssd: A unified framework for self-supervised outlier detection.
arXiv preprint arXiv:2103.12051
(2021)
\end{botherref}
\endbibitem

\bibitem[\protect\citeauthoryear{Startups}{2022}]{YOLOv5data}
\begin{botherref}
\oauthor{\bsnm{Startups}, \binits{A.}}:
Vehicle Registration Plates Dataset.
Roboflow Universe \url{https://universe.roboflow.com/augmented-startups/vehicle-registration-plates-trudk}
(2022)
\end{botherref}
\endbibitem

\bibitem[\protect\citeauthoryear{Marti and Bunke}{2002}]{iam}
\begin{barticle}
\bauthor{\bsnm{Marti}, \binits{U.-V.}},
\bauthor{\bsnm{Bunke}, \binits{H.}}:
\batitle{The iam-database: An english sentence database for offline handwriting recognition}.
\bjtitle{International Journal on Document Analysis and Recognition}
\bvolume{5},
\bfpage{39}--\blpage{46}
(\byear{2002})
\doiurl{10.1007/s100320200071}
\end{barticle}
\endbibitem

\bibitem[\protect\citeauthoryear{Smith}{2015}]{tesseractcode}
\begin{botherref}
\oauthor{\bsnm{Smith}, \binits{R.}}:
Tesseract Open Source OCR Engine.
\url{https://github.com/tesseract-ocr/tesseract}
(2015)
\end{botherref}
\endbibitem

\bibitem[\protect\citeauthoryear{Kay}{2007}]{tesseractpaper}
\begin{barticle}
\bauthor{\bsnm{Kay}, \binits{A.}}:
\batitle{Tesseract: an open-source optical character recognition engine}.
\bjtitle{Linux J.}
\bvolume{2007}(\bissue{159}),
\bfpage{2}
(\byear{2007})
\end{barticle}
\endbibitem

\bibitem[\protect\citeauthoryear{Krause et~al.}{2013}]{Krause2013stanfrordcars}
\begin{bchapter}
\bauthor{\bsnm{Krause}, \binits{J.}},
\bauthor{\bsnm{Deng}, \binits{J.}},
\bauthor{\bsnm{Stark}, \binits{M.}},
\bauthor{\bsnm{Fei-Fei}, \binits{L.}}:
\bctitle{Collecting a large-scale dataset of fine-grained cars}.
In: \bbtitle{Proceedings of the IEEE International Conference on Computer Vision Workshops}
(\byear{2013}).
\bcomment{\url{https://api.semanticscholar.org/CorpusID:16632981}}
\end{bchapter}
\endbibitem

\end{thebibliography}


\end{document}